\documentclass[11pt,a4paper]{article}

\usepackage[a4paper,margin=1in]{geometry}
\usepackage{graphicx}
\usepackage[numbers]{natbib}
\usepackage{authblk}
\usepackage[T1]{fontenc}
\usepackage[utf8]{inputenc}
\usepackage{url}
\usepackage{booktabs}
\usepackage{nicefrac}
\usepackage[protrusion=true,expansion=false]{microtype}
\usepackage{xcolor}
\usepackage{subcaption}
\usepackage{multirow}
\usepackage{amsmath,amssymb,amsfonts}
\usepackage{amsthm}
\usepackage{mathrsfs}
\usepackage{enumitem}
\usepackage{float}
\usepackage[section]{placeins}

\usepackage{algorithm}
\usepackage{algorithmicx}
\usepackage{algpseudocode}
\graphicspath{{pics/}{./}}

\newtheorem{theorem}{Theorem}
\newtheorem{proposition}[theorem]{Proposition}
\newtheorem{lemma}{Lemma}

\newtheorem{definition}{Definition}

\newcommand{\R}{\mathbb{R}}
\newcommand{\cN}{\mathcal{N}}

\newcommand{\cK}{\mathcal{K}}
\newcommand{\cD}{\mathcal{D}}

\newcommand{\norm}[1]{\left\|#1\right\|_w}

\newcommand{\hats}{\hat{s}}
\newcommand{\T}{\mathcal{T}}
\newcommand{\cG}{\mathcal{G}}

\usepackage[hidelinks]{hyperref}
\hypersetup{
  pdftitle={TEMPO: Regime-Aware Operator Learning with Locally Adapted POD Bases},
  pdfauthor={Rodion Akinzhala, Alexander Terentyev, Alexey Golubev, Yury Maximov}
}

\begin{document}
\title{TEMPO: Regime-Aware Operator Learning with Locally Adapted POD Bases}
\author{Rodion Akinzhala, Alexander Terentyev, Alexey Golubev, Yury Maximov}
\date{}
\maketitle
\vspace{-1.0cm}

\begin{abstract}

Operator learning approximates the solution map of a parametric partial differential equation from data, enabling fast prediction at new parameter values without repeated simulation. The dominant approach fits a single global low-rank basis, typically a proper orthogonal decomposition (POD), to all solution snapshots. This fails when the snapshots span qualitatively different solution shapes, from smooth diffusive profiles to near-discontinuous shocks.

We propose TEMPO (Trajectory Expectation-Maximization Mixture of POD Operators), a two-phase framework for regime-aware operator learning. In the offline phase, TEMPO recovers latent regimes via expectation-maximization, fitting a dedicated linear or nonlinear basis to each discovered regime.  In the online phase, a gating network and per-regime branch networks are trained jointly, routing new inputs to the regimes at inference.

{%
We test TEMPO on Burgers' flow, Sod shock tubes, Coanda flow, {Darcy flow, }Navier--Stokes flow and the heterogeneous Helmholtz equation. With 64 total POD modes, TEMPO gives {} 18.61\% test error on Burgers',  {compared with 19.61\% for global POD}. It also gives lower errors on Coanda and Helmholtz{}. {}
\par}
\end{abstract}

\noindent\textbf{Keywords:} operator learning; parametric PDE; regime discovery;\\
proper orthogonal decomposition; expectation-maximization; neural operator

\section{Introduction}\label{sec:intro}

Reduced-order modeling of parametric partial differential equations (PDEs) constructs low-rank bases, classically by proper orthogonal decomposition (POD), from solution snapshots to enable fast prediction at new parameter values without re-solving the governing equations~\cite{Hesthaven2016, Kovachki2024}.
{One such approach is} POD-DeepONet~\cite{Lu2022}, a deep
operator network (DeepONet)~\cite{Lu2019} whose trunk is a fixed POD basis.
The basis is computed once from the full snapshot ensemble, and a branch
network predicts the expansion coefficients for unseen inputs{~\cite{Lu2022}}.
This approach succeeds when solutions lie on a smooth low-dimensional manifold.
It fails when the snapshots span qualitatively different solution shapes, from
smooth diffusive profiles to near-discontinuous shocks.
 {A single global basis must represent different solution regimes with the same set of modes.}
{%
Shifted POD uses co-moving frames~\cite{Reiss2018}, while
Tiba et al.~\cite{Tiba2026} interpolate local subspaces and
update them from new high-fidelity snapshots.
TEMPO uses a collection of local space--time bases that remain fixed during prediction.
}

Partitioning the solution or parameter space and fitting local bases to each region is the classical remedy in reduced-order modeling~\cite{Amsallem2012, Hess2019}.
Daniel et al.~\cite{Daniel2020} extend this to deep learning by clustering solution subspaces on the Grassmann manifold and training a dictionary selector {for} inference.
These approaches reduce approximation error on multi-modal problems.
{In Daniel et al.'s approach, the partition is fixed before the local bases are fitted, rather than updated using their reconstruction errors.}
A later work by the same group~\cite{Daniel2022} {introduces} an a
priori efficiency criterion.
{Kang et al.~\cite{Kang2022} use soft clustering to construct local POD models. A separate classifier predicts the weights for combining their outputs at a new input. TEMPO updates regime assignments and local bases together using reconstruction errors, then jointly trains the gate and coefficient networks.}
Joint inference of a partition and of local linear subspaces is classical in
statistics: mixtures of probabilistic principal component
analyzers~\cite{Tipping1999} and mixtures of factor
analyzers~\cite{Ghahramani1996} fit both by expectation-maximization (EM).

A second line of work enriches the basis while retaining a single global model.
Sharma and Shankar~\cite{Sharma2025} ensemble several distinct trunk networks in
a DeepONet: a standard trunk, a partition-of-unity mixture-of-experts trunk, and
a POD trunk.
The experts are gated over the spatial domain, so the mixture is over location
rather than over solution regimes.
 {Nonlinear reduced-order modeling (ROM)} using deep autoencoders~\cite{Lee2020} significantly outperforms the optimal linear subspace on advection-dominated problems.
NeuralPOD~\cite{Mou2026} pursues this direction for operator learning, learning nonlinear mode functions by sequential residual minimization.
These approaches improve representational capacity but treat the full ensemble jointly, without identifying distinct solution regimes.

Neural operators such as the Fourier neural operator (FNO)~\cite{Li2021} and the convolutional neural operator (CNO)~\cite{Raonic2023} learn solution operators as discretization-invariant functional maps, without explicit {trajectory POD bases}. {They do not explicitly identify solution regimes.}

We propose \textbf{TEMPO} (\textit{Trajectory Expectation-Maximization Mixture of POD Operators}),
 {which jointly learns regime assignments and local trajectory bases, then predicts regime weights and expansion coefficients from PDE inputs.}
TEMPO models the trajectory ensemble as a Gaussian mixture and  {estimates regime assignments} via EM~\cite{Dempster1977}.
The E-step assigns each trajectory to regimes based on reconstruction quality; the M-step fits a dedicated basis to each regime.

{}

\textbf{Contributions:}
\begin{itemize}[leftmargin=1.5em]
  \item \textbf{TEMPO framework.} A two-phase mixture-of-operators for
    parametric PDEs: an offline EM that recovers $M$ latent regimes from
    unlabeled snapshots, or from the observed parameter, and fits a dedicated
    orthogonal (POD) or nonlinear (NeuralPOD) basis to each; an online phase where a gating network (GatingNet) and $M$
    branch networks (BranchNets) are trained with a  {} {Kullback--Leibler (KL)} penalty that anchors routing to the
    offline regime partition.

  \item \textbf{Adaptive basis update.} A distribution-shift criterion that
    monitors moment changes in regime composition across EM iterations and
    selects per-regime between full retraining, incremental mode
    addition/pruning, or skipping.

  \item \textbf{Approximation bound.}
    
    {Optimal rank-$K$ bases fitted to each regime ($MK$ modes in total)
    never increase offline reconstruction error relative to a global rank-$K$ basis,
    for any responsibilities (Proposition~\ref{prop:advantage}).}

  \item \textbf{Experiments.} 

{%
 {Tests on six PDE benchmarks} include POD comparisons at equal total mode counts and diagnostics of reconstruction error, prediction error and inference cost.
\par}
\end{itemize}

\section{Problem Statement}\label{sec:problem}

Let $\Omega \subset \R^d$ be a bounded spatial domain and $\T \subset \R_{\geq 0}$ a time interval.

\begin{figure}[!ht]
  \centering
  \includegraphics[width=0.45\textwidth]{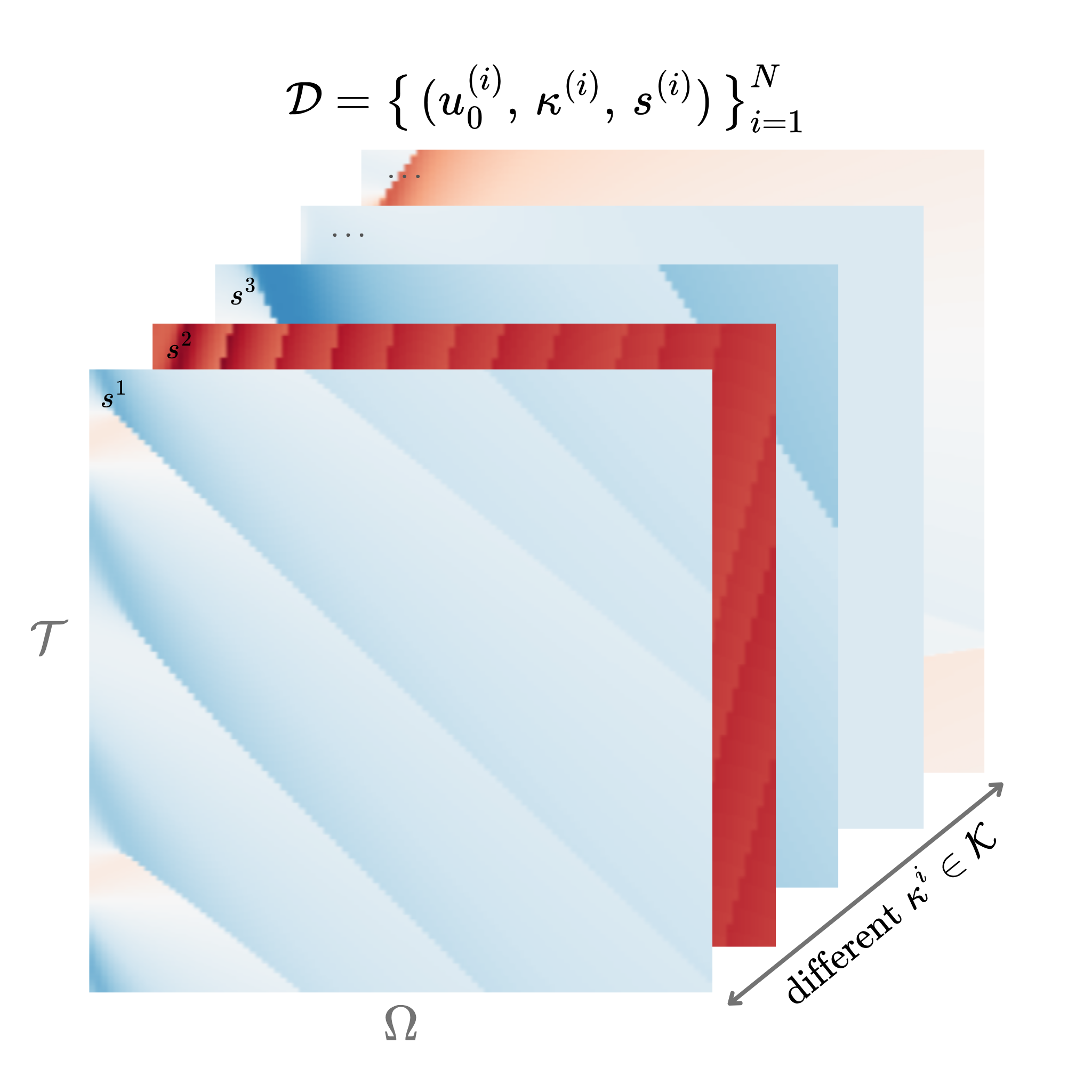}
    \caption{Training dataset $\cD$: solution trajectories $s^{(i)}$ on
$\Omega\times\mathcal{T}$ for different $\kappa^{(i)}\in\cK$.}

  \label{fig:dataset}
\end{figure}

We are given a dataset of $N$ solution trajectories (Figure~\ref{fig:dataset}):
\begin{equation}
  \cD = \bigl\{\,(u^{(i)}_0,\,\kappa^{(i)},\,s^{(i)})\,\bigr\}_{i=1}^{N},
  \qquad
  s^{(i)} = s\!\left(\cdot,\,\cdot\,;\,u^{(i)}_0,\kappa^{(i)}\right)
  \in L^2(\Omega\times\mathcal{T}),
  \label{eq:dataset}
\end{equation}
where $u_0^{(i)} \in \mathcal{U} = L^2(\Omega)$ {denotes an input field, such as an initial condition or permeability, and} $\kappa^{(i)} \in \cK$ {collects physical or geometric parameters}.
Each trajectory $s^{(i)}$ is discretized 
{%
into $N_y$ sampled scalar entries at $y_j=(x_j,t_j)$, including all output components%
}, with weighted norm
$\|f\|_w^2 = \sum_j w_j\,{f_j^2}$.
{%
We use fixed uniform weights $w_j=1/N_y$, so the squared residual norm
equals the mean squared error.
}

\begin{definition}[Solution operator]
\label{def:operator}
The \emph{solution operator}
 \[ \cG: \mathcal{U} \times \cK \to L^2(\Omega \times \T) \]
maps each pair $(u_0,\kappa)$ to the full space-time solution:
$\cG(u_0,\kappa) = s(\,\cdot\,,\,\cdot\,;\,u_0,\kappa)$.
\end{definition}

The operator $\cG$ is unknown and expensive to evaluate directly. The \emph{operator learning problem} is to construct $\hat{\cG} \approx \cG$ from $\cD$ alone, generalizing to unseen $(u_0,\kappa) \notin \cD$ without solving the PDE.

\paragraph{Multi-regime assumption}

When the trajectory ensemble contains groups of distinct solution shape, the optimal low-dimensional representation differs from group to group, and no single global basis captures the full solution diversity. We formalize this observation via a latent variable model over the trajectory ensemble.

\begin{definition}[Generative model]
\label{def:generative}
Each trajectory arises from one of $M$ latent regimes. The regime label $z_i \in \{1,\ldots,M\}$ is unobserved and drawn from a categorical prior,
\begin{equation}
  z_i \sim \mathrm{Categorical}(\pi_1,\ldots,\pi_M),
  \qquad \pi_m \geq 0,\quad \textstyle\sum_m \pi_m = 1.
  \label{eq:prior}
\end{equation}
Conditioned on $z_i = m$, the trajectory is distributed as
\begin{equation}
 (s^{(i)}\mid z_i=m)\sim
  \cN\!\bigl(\hats_m^{(i)},\;\sigma^2 W^{-1}\bigr),
  \label{eq:likelihood}
\end{equation}
where $W = \mathrm{diag}(w_1,\ldots,w_{N_y})$ is the quadrature weight matrix, $\sigma^2 > 0$ controls the softness of regime assignments, and $\hats_m^{(i)}$ is the regime-$m$ approximation to $s^{(i)}$.
{%
The expansion coefficients are fitted parameters. 
}
\end{definition}

The generative model leaves $\hats_m^{(i)}$ unspecified. We require only that it admits a \emph{low-rank trajectory decomposition}:
\begin{equation}
  \hats_m^{(i)}(y)
  \;=\; \phi_0^{(m)}(y)
  + \sum_{k=1}^{K_m}
    \lambda_k^{(m,i)}\,\phi_k^{(m)}(y),
  \qquad y = (x,t) \in \Omega\times\mathcal{T},
  \label{eq:separable}
\end{equation}
where $\phi_k^{(m)} : \Omega\times\mathcal{T}\to\mathbb{R}$ are spatiotemporal
mode functions and $\lambda_k^{(m,i)} \in \mathbb{R}$ a scalar amplitude
for trajectory~$i$ on mode~$k$.
The linear case recovers POD-DeepONet~\cite{Lu2022}. The nonlinear case uses network-parametrized modes~\cite{Mou2026}. Both are studied in this work.

\begin{definition}[Regime approximator]
\label{def:approx}
The parameters of regime $m$ are the mode functions
$\Phi_m = \{\phi_k^{(m)}\}_{k=0}^{K_m}$, with $\phi_0^{(m)}$ the regime mean
field, and the amplitudes
$\Lambda_m = \{\lambda_k^{(m,i)}\}_{k=1,\,i=1}^{K_m,\,N}$.
The mode count $K_m \leq K_{\max}$ is adaptive.
\end{definition}

The learning problem decomposes into two phases: discovering the latent regime structure from $\cD$ \emph{offline}, then training a deployable operator for fast inference \emph{online}.

\paragraph{Phase 1: Offline regime discovery}

Since the regime labels $z_i$ in~\eqref{eq:likelihood} are unobserved,
we  {fit} the bases $\{\Phi_m\}${, coefficients $\Lambda$}
and mixture weights $\pi$ by maximizing the observed-data log-likelihood at
fixed $\sigma^2$:

\begin{equation}
  {\max_{\Phi,\,\Lambda,\,\pi}}\;
  \sum_{i=1}^N\log\sum_{m=1}^M
  \frac{\pi_m{h_{im}}\det(W)^{1/2}}{(2\pi\sigma^2)^{N_y/2}}
  \exp\!\left(-\frac{\|s^{(i)}-\hats_m^{(i)}\|_w^2}{2\sigma^2}\right).
  \label{eq:mle}
\end{equation}

{%
Here, $h_{im}=1$ unless a positive scalar parameter $\kappa$ guides assignments. In that case,
$h_{im}=\cN(\log\kappa^{(i)};c_m,T^2/\rho)$ is a normalized Gaussian density,
with centers $c_m$, width $T>0$ and weight $\rho>0$ fixed at initialization.
This is the joint likelihood of the trajectories and $\log\kappa$,
assumed independent within each regime. 
}

\paragraph{Phase 2: Online operator learning}

With bases $\{\Phi_m^*\}$ and responsibilities $\{\gamma_{im}^*\}$ frozen,
we train a gating network (\emph{GatingNet}) and $M$ branch networks (\emph{BranchNets}) that map a new input
$(u_0, \kappa)$ to a weighted combination of local regime predictions.
The precise architecture and loss are given in Section~\ref{sec:online}.
The quality of the final operator is measured by the mean relative $L^2$ error:
\begin{equation}
  \varepsilon \;=\; \frac{1}{N_\mathrm{test}}
  \sum_{i=1}^{N_\mathrm{test}}
  \frac{\norm{s^{(i)} - \hat{s}(\cdot;\,u_0^{(i)},\kappa^{(i)})}}
       {\norm{s^{(i)}}} .
  \label{eq:metric}
\end{equation}

\section{Method}\label{sec:method}

{TEMPO initializes regime assignments from a Gaussian mixture model (GMM) fitted to global POD coefficients or from physical parameters, then fits the local bases. It alternates reconstruction-based assignment updates and basis fitting (Algorithm~\ref{alg:tempo}).}

All methods studied in this work share the two-phase structure of
Section~\ref{sec:problem} and differ along two independent axes:

\begin{itemize}
  \item \textbf{Basis type.}
    A \emph{linear} (POD) basis uses the singular value decomposition (SVD) of the (weighted) snapshot
    matrix; a \emph{nonlinear} (NeuralPOD) basis learns spatiotemporal
    mode networks by sequential residual minimization.
  \item \textbf{Number of regimes.}
    A \emph{single-regime} operator fits one global basis to all
    snapshots; a \emph{multi-regime} operator (TEMPO) discovers $M$
    latent regimes via EM and assigns a dedicated basis to each.
\end{itemize}

\noindent
This gives four combinations:
POD-DeepONet~\cite{Lu2022} (linear, global),
NeuralPOD-DeepONet (nonlinear, global, our extension of~\cite{Mou2026}),
TEMPO(POD) (linear, $M$ regimes), and
TEMPO(NeuralPOD) (nonlinear, $M$ regimes).

\subsection{Basis Representations}\label{sec:bases}

\subsubsection{POD Basis}

POD~\cite{Berkooz1993} extracts a hierarchy of orthogonal modes that capture the maximum energy from the snapshot ensemble.
Let $\phi_0 := \bar{s} = \tfrac{1}{N}\sum_i s^{(i)}$ be the mean field
and $\tilde{S} \in \mathbb{R}^{N_y \times N}$ the matrix with columns
$\tilde{s}^{(i)} = s^{(i)} - \phi_0$.
The rank-$K$ truncated SVD $\tilde{S} \approx U_K \Sigma_K V_K^\top$
yields orthonormal modes $\{\phi_k\}_{k=1}^K$ (columns of $U_K$) and
coefficients $\lambda_k^{(i)} = [\Sigma_K V_K^\top]_{ki}${.}
{%
Here, $U_K$ and $V_K$ contain the first $K$ left and right singular vectors,
and $\Sigma_K$ is diagonal with the corresponding singular values.
}
The approximation
\begin{equation}
  \hat{s}^{(i)} = \phi_0 + \sum_{k=1}^{K} \lambda_k^{(i)}\,\phi_k
  \label{eq:pod_approx}
\end{equation}
is optimal in the $L^2$ sense among all rank-$K$ affine representations.

For TEMPO(POD), the M-step~\eqref{eq:mstep_phi} is solved analytically\ {at fixed rank}:
set $\phi_0^{(m)} := \bar{s}_m = \frac{1}{N_m}\sum_i \gamma_{im} s^{(i)}$,
where $N_m = \sum_i \gamma_{im}$, and apply SVD to
the weighted centered matrix with columns
$\sqrt{\gamma_{im}}\,(s^{(i)}-\phi_0^{(m)})$.
This is the closed-form linear analogue of Algorithm~\ref{alg:npod}.

\subsubsection{NeuralPOD Basis}\label{sec:neuralpod_basis}

To represent the modes as continuous functions of $y$ rather than as grid
vectors, we parametrize each mode
function $\phi_k : \Omega{\times}\mathcal{T} \to \mathbb{R}$ as a
Fourier feature network~\cite{Tancik2020}; we call the resulting basis a \emph{Fourier NeuralPOD} basis:
\begin{equation}
  \phi_k(y) \;=\; f_{\theta_k}\!\bigl(\psi(y)\bigr),
  \qquad
  \psi(y) \;=\;
  \bigl[\sin(2\pi By),\;\cos(2\pi By)\bigr],
  \label{eq:ffn}
\end{equation}
where $B \in \mathbb{R}^{F \times (d+1)}$ is a fixed random matrix drawn
once at initialization and $f_{\theta_k} : \mathbb{R}^{2F} \to \mathbb{R}$
is a small multilayer perceptron (MLP) with Tanh activations.
Plain MLPs exhibit spectral bias~\cite{Rahaman2019}; the Fourier encoding
reduces it.
For multi-scale coverage, the rows of $B$ are partitioned evenly across
$L$ frequency scales $\rho_1 < \cdots < \rho_L$: the $\ell$-th block is drawn from
$\mathcal{N}(0,\,\rho_\ell^2 I)$.
The mean field $\phi_0$ uses the same architecture.

The scalar amplitudes $\lambda_k^{(i)} \in \mathbb{R}$ are direct
learnable parameters, one per training trajectory, optimized jointly
with $\phi_k$.

Modes are learned by \emph{sequential residual minimization}~\cite{Mou2026}.
Let $r^{(0,i)} = s^{(i)} - \phi_0$, where $\phi_0$ is fitted to the snapshot mean $\bar{s}$ defined above.
For each subsequent mode $k \ge 1$, given the residual
\begin{equation}
  r^{(k-1,i)}(y)
  = s^{(i)}(y)
  - \phi_0(y)
  - \sum_{\ell=1}^{k-1}\lambda_\ell^{(i)}\,\phi_\ell(y),
  \label{eq:residual_global}
\end{equation}
we solve
\begin{equation}
  \min_{\phi_k,\,\{\lambda_k^{(i)}\}}
  \sum_{i=1}^{N} \sum_j w_j
  \Bigl(\lambda_k^{(i)}\,\phi_k(y_j) - r_j^{(k-1,i)}\Bigr)^{\!2}
  \label{eq:neuralpod_loss}
\end{equation}
(the global case with uniform weights; the $\gamma$-weighted
generalization used in TEMPO is Algorithm~\ref{alg:npod}).
Mode addition stops when the residual falls below fraction $\tau$ of
the initial variance,
\begin{equation}
  \sum_{i=1}^N \|r^{(k,i)}\|_w^2
  \;<\;
  \tau \cdot \sum_{i=1}^N \|s^{(i)} - \bar{s}\|_w^2,
  \label{eq:npod_tol}
\end{equation}
or $k = K_{\max}$.
 {}

{For TEMPO(NeuralPOD), we {may jointly refine} each regime's sequentially fitted mean and modes at fixed rank, alternating gradient updates of the basis networks with least-squares updates of the training coefficients. The bases are then frozen for {online training}.}

\subsection{Single-Regime Operators}\label{sec:single}

\subsubsection{POD-DeepONet}

POD-DeepONet~\cite{Lu2022} replaces the learned trunk network of
DeepONet~\cite{Lu2019} with the fixed POD basis.
After computing the global modes $\{\phi_k\}_{k=1}^K$ and training
coefficients $\{{\boldsymbol{\lambda}^{(i)}}\}$ offline, a branch network
$\mathcal{B}_\theta : \mathbb{R}^{m_s + d_\kappa} \to \mathbb{R}^K$,
where $u_0$ is sampled at $m_s$ sensor points and
$\kappa \in \mathbb{R}^{d_\kappa}$,
{is trained by minimizing coefficient error or field prediction error. The coefficient-regression objective is:}
\begin{equation}
  \mathcal{L}_{\mathrm{branch}}
  = \frac{1}{N}\sum_{i=1}^{N}
    \bigl\|\mathcal{B}_\theta(u_0^{(i)},\,\kappa^{(i)})
    - {\boldsymbol{\lambda}^{(i)}}\bigr\|^2.
  \label{eq:pod_branch_loss}
\end{equation}
Lu et al. train the branch through the output loss.  {} {For an orthonormal basis, coefficient regression and the absolute squared field loss have the same minimizer.}
The prediction at any $(x,t) \in \Omega \times \mathcal{T}$ is
\begin{equation}
  \hat{s}(x,t;\,u_0,\kappa)
  = \phi_0(x,t)
  + \sum_{k=1}^{K}
    \mathcal{B}_{\theta,k}(u_0,\kappa)\,\phi_k(x,t).
  \label{eq:pod_predict}
\end{equation}

\subsubsection{NeuralPOD-DeepONet}\label{sec:neuralpod_deeponet}

Mou et al.~\cite{Mou2026} learn basis functions with networks by sequential
residual minimization and pair them with a DeepONet branch network.
We follow this construction with modes defined on $(x,t)$ jointly and one scalar
amplitude per trajectory.
We call the result \emph{NeuralPOD-DeepONet} and use it as a single-regime
baseline and as the per-regime basis learner within TEMPO(NeuralPOD).

The procedure follows POD-DeepONet with the Fourier NeuralPOD basis in place of POD.
Offline fitting yields coefficients $\boldsymbol{\lambda}^{(i)} \in \mathbb{R}^K$. {Prediction follows~\eqref{eq:pod_predict}.}
At equal mode count $K$ the representation~\eqref{eq:separable} is again
rank-$K$ affine, so POD attains the lowest reconstruction error on the training
grid. The NeuralPOD modes are continuous functions of $y$ and can be evaluated
off the grid, which the POD modes cannot.

\subsection{TEMPO: Offline Regime Discovery}\label{sec:offline}

TEMPO constructs a dedicated basis for each of $M$
latent regimes, recovered via EM on the generative model of
Definition~\ref{def:generative}.

\paragraph{Initialization}
A global POD is computed on all $N$ trajectories; each trajectory is
projected to a $d_0$-dimensional coefficient vector
$\alpha^{(i)} \in \mathbb{R}^{d_0}$.
Running a standard  {%
GMM
} on $\{\alpha^{(i)}\}$ yields initial
responsibilities $\gamma_{im}^{(0)}$, weights $\pi_m^{(0)}$,
and per-regime statistics $(a_m^{(0)}, C_m^{(0)})$.
{Alternatively, Gaussian weights of width $T$ in $\log\kappa$ can initialize the assignments.}
Each regime basis $\Phi_m^{(0)}$ is then fitted to the
responsibility-weighted ensemble: for TEMPO(NeuralPOD) via
\textsc{NeuralBasis} (Algorithm~\ref{alg:npod}); for TEMPO(POD)
via the weighted SVD of Section~\ref{sec:bases}.
The scale $\sigma^2$ sets how soft the assignments are. It is fixed once
from the initial reconstructions rather than tuned:
\begin{equation}
  \sigma^2 \;\leftarrow\;
  \frac{1}{2N}\sum_{i=1}^N\sum_{m=1}^M
  \gamma_{im}^{(0)}\,\bigl\|s^{(i)} - \hat{s}_m^{(i)}\bigr\|_w^2.
  \label{eq:sigma2_calib}
\end{equation}
This makes the  {reconstruction term}
in~\eqref{eq:estep} of order one for a typical trajectory.
$\sigma^2$ is held fixed throughout EM{.}

\paragraph{E-step}
At iteration $q$, the responsibility of trajectory $i$ for regime $m$
is updated by Bayes' rule on the Gaussian likelihood~\eqref{eq:likelihood}:
\begin{equation}
  \gamma_{im}^{(q)}
  = \frac{
    \pi_m^{(q-1)}\,{h_{im}}\,
    \exp\!\Bigl(-\dfrac{\|s^{(i)}-\hat{s}_m^{(i)}\|_w^2}{2\sigma^2}\Bigr)
  }{
    \sum_{j=1}^{M}
    \pi_j^{(q-1)}\,{h_{ij}}\,
    \exp\!\Bigl(-\dfrac{\|s^{(i)}-\hat{s}_j^{(i)}\|_w^2}{2\sigma^2}\Bigr)
  },
  \label{eq:estep}
\end{equation}
where $\hat{s}_m^{(i)}$ is evaluated via~\eqref{eq:separable}
using $\Phi_m^{(q-1)}$ and $\Lambda_m^{(q-1)}$.
This reconstruction-based form of the responsibilities follows Hinton et
al.~\cite{Hinton1997}.
 {Assignments depend on the current local reconstructions, rather than on a fixed feature representation.}

\paragraph{M-step}
With responsibilities fixed, the M-step gives a closed-form update
for the mixture weights,
\begin{equation}
  \pi_m^{(q)} \leftarrow \frac{N_m^{(q)}}{N},
  \qquad N_m^{(q)} = \sum_{i=1}^{N}\gamma_{im}^{(q)},
  \label{eq:mstep_pi}
\end{equation}
and a weighted reconstruction problem for each basis:
\begin{equation}
  \min_{\Phi_m,\,\Lambda_m}
  \sum_{i=1}^{N} \gamma_{im}^{(q)}\,\|s^{(i)} - \hat{s}_m^{(i)}\|_w^2,
  \label{eq:mstep_phi}
\end{equation}
{\raggedright
solved by weighted SVD for TEMPO(POD) or
\textsc{NeuralBasis} (Algorithm~\ref{alg:npod}) for
TEMPO(NeuralPOD).
\par}

{Adaptive or approximate basis updates can decrease the likelihood between iterations. The relative-error variant normalizes squared reconstruction errors by squared trajectory norms throughout offline fitting and is treated as a reconstruction-based pseudo-likelihood.}

\begin{proposition}[{Optimal offline reconstruction}]\label{prop:advantage}
For any $\{\gamma_{im}\}\geq 0$ with $\sum_m\gamma_{im}=1$, any $M\geq 1$, any $\;{1\leq K\leq N_y}$:
\[
  \varepsilon_{\mathrm{TEMPO}}^K \;\leq\; \varepsilon_{\mathrm{global}}^K,
\]
where
\begin{align*}
  \varepsilon_{\mathrm{global}}^K &= \min_{\substack{P\,\mathrm{rank\text{-}}K\\\mu}}\sum_i\|(I{-}P)(s^{(i)}{-}\mu)\|_w^2, \\
  \varepsilon_{\mathrm{TEMPO}}^K  &= \sum_m\min_{\substack{P_m\,\mathrm{rank\text{-}}K\\\mu_m}}\sum_i\gamma_{im}\|(I{-}P_m)(s^{(i)}{-}\mu_m)\|_w^2
\end{align*}
are the global and regime-wise optimal rank-$K$ reconstruction errors.
{%
Here, $P$ and $P_m$ are $W$-orthogonal projectors.
}
Proof: see \ref{app:proof}.
\end{proposition}

The $M$ regime bases together store $MK$ modes, against $K$ for the global
basis.
{%
The bound concerns optimal training reconstruction; online accuracy also
depends on the predicted weights and coefficients.
}
For hard partitions the inequality is standard~\cite{Daniel2022}.

\begin{figure}[H]
\centering
\includegraphics[width=\textwidth]{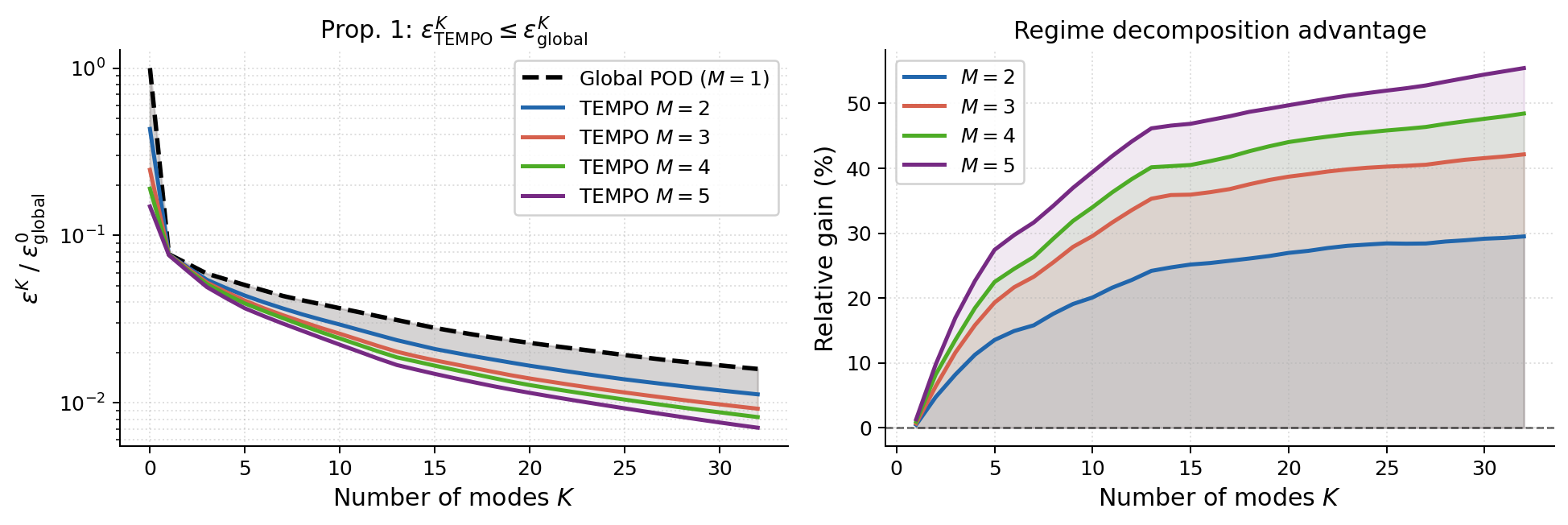}
\caption[TEMPO(POD) reconstruction with fixed assignments]{
{} {} {Offline reconstruction using the TEMPO(POD) representation: $M$ regimes with $K$ modes each ($MK$ total), versus $K$ global POD modes. Left: normalized error; right: relative error reduction.}}
\label{fig:theorem}
\end{figure}

Figure~\ref{fig:theorem}  {illustrates}
Proposition~\ref{prop:advantage} on 1D Burgers' ($N=1{,}500$,
$\nu\in\{0.001,0.1,1.0\}$) with {fixed} $k$-means regime
assignments.

\paragraph{Adaptive basis update}
Retraining all $M$ NeuralPOD bases at every EM iteration is
computationally prohibitive.
We introduce a distribution-shift criterion to determine whether each
basis requires updating.
Let $\alpha^{(i)}$ denote the global POD projection of $s^{(i)}$,
computed once during initialization and fixed thereafter.
The responsibility-weighted first and second moments of regime $m$ are
\begin{align}
  a_m^{(q)}
  &= \frac{1}{N_m^{(q)}}\sum_{i=1}^N
    \gamma_{im}^{(q)}\,\alpha^{(i)},
  \label{eq:mu}\\
  C_m^{(q)}
  &= \frac{1}{N_m^{(q)}}\sum_{i=1}^N
    \gamma_{im}^{(q)}\,
    (\alpha^{(i)}-a_m^{(q)})(\alpha^{(i)}-a_m^{(q)})^\top.
  \label{eq:Sigma}
\end{align}
The distribution-shift criterion measures the relative change in these
moments between consecutive iterations:
\begin{equation}
  \Delta_m^{(q)}
  = \frac{\|a_m^{(q)} - a_m^{(q-1)}\|_2}
         {\|a_m^{(q-1)}\|_2 + \delta}
  + \frac{\|C_m^{(q)} - C_m^{(q-1)}\|_F}
         {\|C_m^{(q-1)}\|_F + \delta},
  \label{eq:Delta}
\end{equation}
where $\delta > 0$ is a small numerical stabilizer.
$\Delta_m^{(q)}$ measures how much the composition of regime $m$ has changed,
not just the volume of reassignments.
Given thresholds $0 < \theta_s < \theta_l$, regime $m$ is
\textsc{Skip}ped if $\Delta_m^{(q)} < \theta_s$, updated
\textsc{Incremental}ly (add or prune one mode) if
$\theta_s \le \Delta_m^{(q)} < \theta_l$, and fully retrained
from scratch (\textsc{Full rerun}) otherwise.
In the \textsc{Incremental} case, we compute the residual of the
current basis under the updated responsibilities $\gamma^{(q)}$, add
a new mode if $\sum_i \gamma_{im}^{(q)}\|r_m^{(K_m,i)}\|_w^2 \ge \tau\cdot\sum_i \gamma_{im}^{(q)}\|s^{(i)}-\bar{s}_m^{(q)}\|_w^2$,
and prune the last mode if its {mean} weighted contribution falls below
$\theta_p$.

{%
These updates apply to NeuralPOD, with random initialization for a full
rerun. POD recomputes the weighted SVD whenever a basis update is required.
}

\paragraph[Convergence]{{} {\normalfont\normalsize\itshape Convergence}}

{}
EM terminates when $\max_m \Delta_m^{(q)} < \theta_c${ or the maximum number of iterations is reached.}

The complete offline procedure is given in Algorithm~\ref{alg:tempo}.

\subsection{TEMPO: Online Gated Operator}\label{sec:online}

After Algorithm~\ref{alg:tempo} converges, all regime bases
$\{\Phi_m^*\}$ and responsibilities $\{\gamma_{im}^*\}$ are frozen.
The online phase trains a GatingNet
$G: \mathbb{R}^{m_s + d_\kappa} \to \Delta^{M-1}$
that maps $(u_0, \kappa)$ to regime weights, and $M$ BranchNets
$\mathcal{B}_m: \mathbb{R}^{m_s + d_\kappa} \to \mathbb{R}^{K_m}$
that predict expansion coefficients for each regime.
The TEMPO prediction at query point $y = (x,t) \in \Omega\times\mathcal{T}$ is
\begin{equation}
  \hat{s}(y;\,u_0,\kappa)
  = \sum_{m=1}^{M} g_m(u_0,\kappa)
    \left[
      \phi_0^{(m)}\!(y)
      + \sum_{k=1}^{K_m}
        b_{m,k}(u_0,\kappa)\;
        \phi_k^{(m)}\!(y)
    \right],
  \label{eq:predict}
\end{equation}
where $\mathbf{g} = G(u_0,\kappa)$ and
$\mathbf{b}_m = \mathcal{B}_m(u_0,\kappa)$.
With the NeuralPOD basis the prediction is evaluable at any
$y\in\Omega\times\mathcal{T}$; with the POD basis it is defined on the
training grid.

\paragraph{Training objective}
Writing $g_m^{(i)} = g_m(u_0^{(i)},\kappa^{(i)})$, the online loss is
\begin{equation}
  \mathcal{L}_{\mathrm{online}}
  = \mathcal{L}_{\mathrm{data}}
  + \eta_{\mathrm{KL}}\,\mathcal{L}_{\mathrm{KL}}
  - \eta_{H}\,\mathcal{L}_{\mathrm{ent}}.
  \label{eq:online_loss}
\end{equation}
 {By default,} $\mathcal{L}_{\mathrm{data}} = (1/N)\sum_{i=1}^{N}\|s^{(i)} - \hat{s}(\cdot;\,u_0^{(i)},\kappa^{(i)})\|_w^2$ drives reconstruction accuracy; it is the
mean squared error over 
{the sampled field entries.}
{{}{{} {} {For relative-error training, each squared residual is divided by the squared target norm before averaging.}}}
$\mathcal{L}_{\mathrm{KL}} = (1/N)\sum_{i,m} g_m^{(i)}\log\bigl(g_m^{(i)}/\gamma_{im}^*\bigr)$ 
{penalizes deviations from the offline responsibilities.}
We use the reverse form $\mathrm{KL}(\mathbf{g}^{(i)}\|\boldsymbol{\gamma}_i^*)${.} $\eta_{\mathrm{KL}}$ and
$\eta_H$ balance anchoring against diversity.
$\mathcal{L}_{\mathrm{ent}} = -(1/N)\sum_{i,m} g_m^{(i)}\log g_m^{(i)}$ is the mean entropy of the gating
distribution{. The entropy term encourages weights to be spread across regimes.}
The hyperparameters $\eta_{\mathrm{KL}}, \eta_H \ge 0$ trade
off these objectives.
{} {Loss weights and learning rates may vary during training.}

\section{Computational Experiments}
\label{sec:exp}

\subsection{Experimental Setup}
\label{sec:setup}

{%
 {We consider six benchmarks: Burgers', Euler shock tubes, Coanda flow, Darcy flow, incompressible Navier--Stokes and heterogeneous Helmholtz problems.} Burgers' data are from PDEBench~\cite{Takamoto2022}.
\par}

\paragraph{1D Burgers' equation}
\begin{equation}
  u_t + u\,u_x = \;{\frac{\nu}{\pi}\,u_{xx}},
  \qquad x\in\,{(0,1)},\quad t\in(0,2],
  \label{eq:burgers}
\end{equation}
with periodic boundary conditions and random initial conditions.
Viscosity {parameter} $\nu\in\{0.001,\,0.01,\,0.1,\,1.0\}$ ranges from smooth diffusive
profiles at $\nu{=}1.0$ to near-discontinuous shock structures at
$\nu{=}0.001$.

{%
We retain $1{,}024$ spatial points and $201$ frames, including $t=0$. We use $4{,}000$ training and $1{,}000$ test trajectories at each of $\nu\in\{0.001,0.1,1.0\}$.  {} {A separate generalization test uses $1{,}000$ trajectories at $\nu=0.01$.}
\par}

\paragraph[1D Euler equations]{{1D Euler equations}}
{%
 {We consider Sod-type shock tubes for the one-dimensional Euler equations ($\gamma=1.4$)}~\cite{Sod1978,Peyvan2024}.
\begin{equation}
 \begin{aligned}
 \partial_t U+\partial_x F(U)&=0,\qquad
 E=\frac{p}{\gamma-1}+\frac{1}{2}\rho u^2,\\
 U&=\begin{pmatrix}\rho\\\rho u\\E\end{pmatrix},\qquad
 F(U)=\begin{pmatrix}\rho u\\\rho u^2+p\\u(E+p)\end{pmatrix}.
 \end{aligned}
 \label{eq:sod_euler}
\end{equation}

{The operator maps initial-state parameters and the pressure ratio $p_R/p_L$ to density, velocity and pressure trajectories. The three fields are concatenated into one trajectory vector for joint basis fitting. Targets are exact Riemann solutions on $[0,1]$ up to $t=0.12$. We use $1{,}232$ training trajectories and $1{,}024$ test trajectories from unseen initial states (\ref{app:sod_data}).}
\par}

{
\paragraph{Coanda flow}
 {We use GCA-ROM data for steady incompressible flow through a sudden-expansion channel $\Omega_w$ with inlet half-width $w$}~\cite{pichi2024gcarom,pichiGcaRomRelease}. 
\begin{equation}
 (\mathbf{u}\cdot\nabla)\mathbf{u}-\nu\Delta\mathbf{u}+\nabla p=0,
 \qquad \nabla\cdot\mathbf{u}=0,\qquad \mathbf{x}\in\Omega_w,
 \label{eq:coanda}
\end{equation}

{The operator maps $w$ and viscosity $\nu$ to the velocity field $\mathbf{u}=(u,v)$. We use 217 training cases and 62 test cases on unseen channel geometries (\ref{app:coanda_data}).}
\par}

{%
\paragraph{2D Darcy flow}
\begin{equation}
  -\nabla\cdot\bigl(a(x)\,\nabla u(x)\bigr) = \beta,
  \qquad x\in(0,1)^2,
  \label{eq:darcy}
\end{equation}
with zero Dirichlet boundary conditions.
Here $a(x)$ is a spatially varying permeability field, and $\beta$ is the
constant source term. We use PDEBench permeability fields~\cite{Takamoto2022}
and compute stationary pressures with a conservative finite-volume solver
on a $128\times128$ uniform grid.
 {We use $7{,}000$ training and $1{,}000$ test permeability fields.}

{All models predict pressure from permeability at $\beta=1$ and apply the exact scaling $u(a,\beta)=\beta u(a,1)$. TEMPO therefore learns regimes from pressure variations caused by permeability (\ref{app:darcy_data}).}
\par}

\paragraph{2D incompressible Navier--Stokes}
\begin{equation}
  \partial_t \omega + \mathbf{u}\cdot\nabla\omega
  = \;{\nu_{\mathrm{Re}}}\,\Delta\omega + f,
  \qquad \mathbf{x}\in\,{[0,2\pi]^2},\; t\in(0,2],
  \label{eq:ns}
\end{equation}
with periodic boundary conditions, \ {signed vorticity $\omega=\partial_{x_2}u-\partial_{x_1}v=\Delta\psi$},
divergence-free $\mathbf{u}$ recovered via the stream function\ {as $\mathbf{u}=(\partial_{x_2}\psi,-\partial_{x_1}\psi)$},
and \ {forcing}
\[
  f(\mathbf{x})={0.1\bigl(\sin}{(x_1{+}x_2)+\cos}{(x_1{+}x_2)\bigr).}
\]

Reynolds number $\mathrm{Re}\in\{100,\,1000,\,3600,\,10000\}$ ranges from
smooth laminar flow to convection-dominated flow{, with viscosity $\nu_{\mathrm{Re}}\simeq\mathrm{Re}^{-1}$}.
Each value provides $N{=}5{,}000$ trajectories on a $64{\times}64$ periodic
grid ($20{,}000$ total; $4{,}000$/$1{,}000$ train/test);
see \ref{app:ns_data}. 
The operator maps {$(\mathbf{u}_0,\mathrm{Re})$} to the sequence
{$(\mathbf{u}_t)_{t=0}^{20}$} with $\Delta t{=}0.1$.

{
\paragraph{Heterogeneous Helmholtz equation}
{}
We extend the benchmark of de Hoop et al.~\cite{dehoop2022cost,dehoop2022operatordata}  {to predict the wave field $u$ from the wavespeed field $c$ and frequency $\omega\in\{600,800,1000,1200\}$.} 
\begin{equation}
 -\Delta u-\left(\frac{\omega}{c(\mathbf{x})}\right)^2u=0,
 \qquad \mathbf{x}\in(0,1)^2,
 \label{eq:helmholtz}
\end{equation}
{}
 {We impose localized Neumann excitation. We use } {4,000 training and 800 test solutions}\ {on a $101\times101$ grid. Wavespeed fields are disjoint across splits (\ref{app:helmholtz_data}).} 
\par}

\paragraph{Methods}
We compare DeepONet~\cite{Lu2019}, POD-DeepONet~\cite{Lu2022},
FNO~\cite{Li2021}, and CNO~\cite{Raonic2023}, all trained jointly on all
parameter values via a single global model.

{We also compare with NeuralPOD-DeepONet~\cite{Mou2026}, implemented with a learned Fourier basis and a branch network. TEMPO(POD) and TEMPO(NeuralPOD) use local bases learned via EM and a gating network.} 
{Fixed GMM + POD fits local bases using fixed GMM responsibilities, then trains the gating and branch networks without further assignment updates.}
{%
 {Per-parameter specialists are reported in Table~\ref{tab:revision_burgers_original}.} FNO and CNO are compared on Sod, Coanda, Helmholtz and Navier--Stokes.
\par}

All methods implemented by the authors.

{}
{\raggedright
Exact architectures, training schedules, and all hyperparameter values
$(\eta_{\mathrm{KL}}, \eta_H, \theta_s, \theta_l, \theta_p, \theta_c, \tau, K_{\max}, d_0, m_s)$
are available in the accompanying code repository.
Evaluation: mean relative $L^2$ error~\eqref{eq:metric} per parameter value.
Inference times measured on a single NVIDIA RTX A5000 GPU, per sample.
\par}

\subsection{Main Results}
\label{sec:results}

\begin{figure}[!htbp]
\centering
\includegraphics[width=\textwidth]{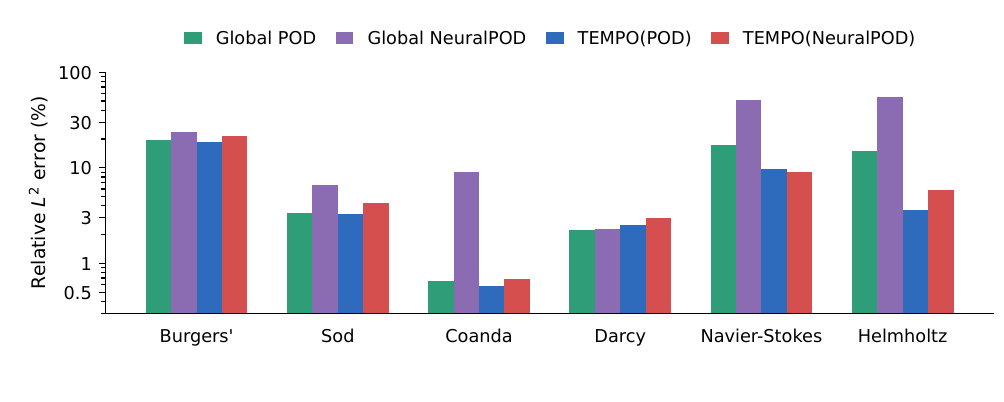}
\caption{{Mean relative $L^2$ test errors. }}
\label{fig:joint_benchmarks}
\end{figure}

{Figure~\ref{fig:joint_benchmarks} summarizes the joint-model errors.}

\paragraph{1D Burgers' equation}

{%
TEMPO(POD) with two 32-mode spaces reaches {} {18.61\%} test error,  {compared with {19.61\%} for global POD64} (Table~\ref{tab:revision_burgers_original}). DeepONet gives {35.84\%}. TEMPO(NeuralPOD) gives {} {} {{21.53\%}, versus {23.85\%} for global NeuralPOD}; they use 48 and 32 modes, respectively.
\par}

{%
\paragraph[Sod shock tubes]{{Sod shock tubes}}
\sloppy
TEMPO(POD) reaches {} {} {{3.27\%}} mean test error,  {} {compared with {3.37\%} for global POD48}. FNO and CNO give {5.78\%} and {4.52\%}; TEMPO(NeuralPOD) gives {} {} {{4.24\%} (Table~\ref{tab:revision_sod})}.

\paragraph{Coanda flow}
TEMPO(POD) gives {0.59\%} mean error,  {compared with {0.65\%} for global POD24} at the same total mode count. TEMPO(NeuralPOD) gives {} {} {{0.68\%}.}  {Tests use two withheld geometries} (Table~\ref{tab:revision_coanda}).
\par}

{%
\paragraph{2D Darcy flow}
Table~\ref{tab:revision_darcy} reports relative pressure errors.
With $64$ total modes, TEMPO(POD) gives ${} {2.53\%}$ test error, compared
with ${2.25\%}$ for global POD and ${2.77\%}$ for Fixed GMM + POD.
{}
TEMPO(NeuralPOD) gives ${2.95\%}$, compared with ${2.28\%}$ for global NeuralPOD.
\par}

\paragraph{2D Navier--Stokes}
Table~\ref{tab:ns} gives the cross-Re error matrix{.}  {With four 64-mode bases, TEMPO(POD) and TEMPO(NeuralPOD) reach mean test errors of {} {{9.65\%} and {8.96\%}, respectively}, on the same $4{,}000$ trajectories (Table~\ref{tab:ns}).}  FNO and CNO are more accurate on this benchmark.

{%
\raggedright
\paragraph{Heterogeneous Helmholtz equation}
{}
At 64 total modes, TEMPO(POD) gives {} 3.60\% mean relative $L^2$ test error,  {versus 14.87\% for global POD} and 6.48\% for CNO (Table~\ref{tab:revision_helmholtz4000}). {TEMPO(NeuralPOD) gives {5.83\%}.}
\par}

\paragraph{Inference speed}

{%
\sloppy
{}
{TEMPO(POD) evaluates all $M$ branch networks via soft gating, remaining fast relative to FNO and CNO. On Sod, Coanda and Helmholtz, it also achieves lower mean test errors than both baselines. For example, on Coanda, its error is {0.59\%} versus {2.34\%} for CNO, with standard inference times of 0.229 and 8.911\,ms.} Measurements use batch size one on an RTX A5000. Timing protocols and comparisons with global models are given in~\ref{app:matched_inference}.
\par}

\subsection{Regime Discovery}
\label{sec:regime_exp}

{%
TEMPO groups trajectories with a shared local basis, so the number of regimes need not equal the number of sampled parameter values. Similar mean responsibilities do not imply identical trajectory assignments. On Burgers', routing depends on both initial conditions and viscosity; on Coanda, it is associated with flow asymmetry.
}

{%
In Navier--Stokes, the regime means share the spatial harmonic of the common forcing. The local POD bases give similar reconstruction errors on the same trajectories. Each branch predicts its own coefficients, and the gate combines the local predictions using soft weights.
}

\begin{figure}[!htbp]
\centering
\includegraphics[width=\textwidth]{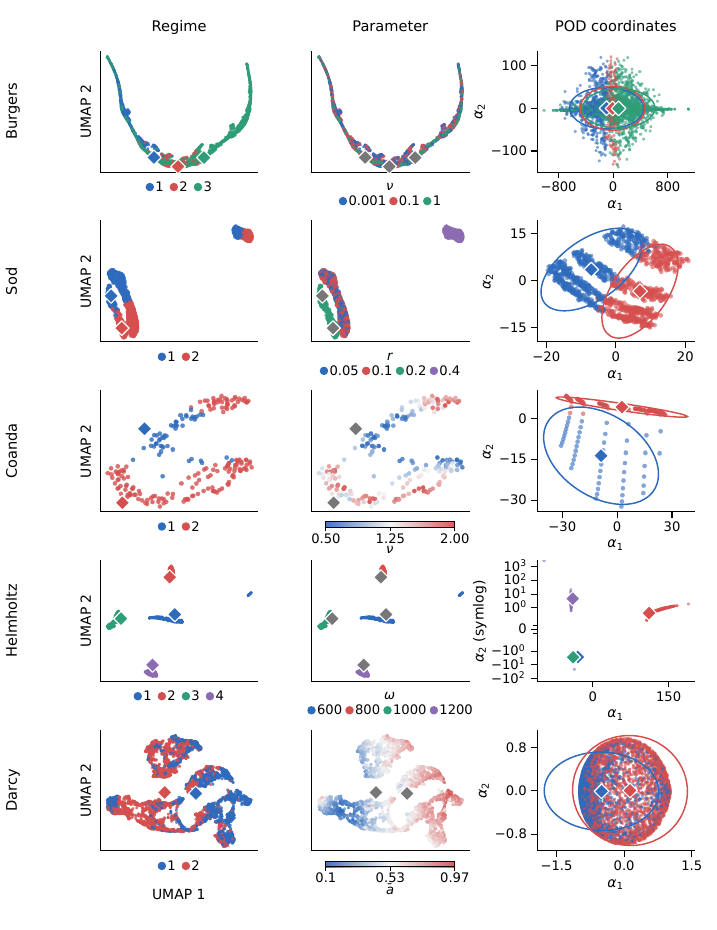}
\caption{{{} {}{} {}} {TEMPO(POD) regimes. UMAP (Uniform Manifold Approximation and Projection) colored by regime (left) and physical quantity (middle); POD coefficients (right). Diamonds mark centers; ellipses show two standard deviations.}}
\label{fig:umap_em}
\label{fig:umap_new_tasks}
\label{fig:umap_all}
\label{fig:helmholtz_regimes}
\label{fig:darcy_regimes}
\end{figure}

{%
Figure~\ref{fig:umap_em} compares the regimes with viscosity $\nu$ (Burgers' and Coanda), pressure ratio $r$ (Sod), frequency $\omega$ (Helmholtz), and spatial mean permeability $\bar a$ (Darcy). The Burgers' row shows a three-regime model. On Helmholtz, the four POD regimes align with the frequencies used for initialization.
}

\subsection{Adaptive Basis Update}
\label{sec:adaptive_exp}

{%
{} {Table~\ref{tab:adaptive} reports update frequencies during {offline EM} in the supplementary three-regime Burgers' models.} {} For POD, incremental updates recompute a weighted SVD, whereas skipping leaves the basis unchanged.
\par}

\subsection{Role of the Physical Parameter}\label{sec:kappa_abl}

{%
{} {Table~\ref{tab:kappa_abl} reports the effect of replacing the parameter by its training mean on Burgers' and Navier--Stokes.}
\par}

{%
 {On Burgers', a classifier identifies viscosity from a 101-sample time series at one spatial point with 96.57\% accuracy. Using its prediction in a four-regime TEMPO(POD) increases reconstruction error over the observed interval by 0.93\% (Table~\ref{tab:viscosity_trace}).}
}

\section{Conclusion}

Operator learning for parametric PDEs usually fits a single global basis to all
snapshots. When the ensemble spans qualitatively different solution shapes, one
basis cannot fit all of them and prediction quality suffers.

TEMPO splits the trajectory ensemble into regimes and gives each regime its own
basis. The partition and the bases are found together: each trajectory gets a
weight in every regime from how well that regime reconstructs it, and each {POD} basis
 {is fitted by} a weighted SVD over all trajectories. {Some} local reduced-order models cluster
first and fit second, so the partition is chosen without reference to the basis
it produces.  {A gating network combines local predictions for new inputs, so one model handles all tested parameter values.}

{%
{A separate 64-mode TEMPO(POD) gives {26.69\%} mean relative $L^2$ error at the unseen Burgers' viscosity $\nu=0.01$, with no training data at that viscosity.}
\par}

{%
 {} {} {}
{At equal total mode counts, TEMPO(POD) improves on global POD on Burgers', Coanda, Navier--Stokes and Helmholtz, with similar accuracy on Sod. It is also faster and more accurate than FNO and CNO on Sod, Coanda and Helmholtz. TEMPO(NeuralPOD) is more accurate than TEMPO(POD) on Navier--Stokes. With exact source scaling, global POD is more accurate on Darcy.}
\par}

{%

\par}

{%
\sloppy
{TEMPO(NeuralPOD) does not consistently outperform TEMPO(POD).} {On Burgers', most of the NeuralPOD error remains after least-squares coefficient fitting.}
\par}

{%
The computational cost of TEMPO grows with the size of the snapshot ensemble
and the number of regimes. Soft gating evaluates all branch networks.
}

{%
Extending the method to higher-dimensional parameter
spaces also requires representative training data. Adaptive sampling could focus
additional simulations on parameter regions with high estimated prediction error.
Future work could adapt local function spaces to the equations and geometry,
using prediction accuracy and cost rather than $L^2$ reconstruction alone.
}

\section*{Data availability}

{ {The 1D Burgers' data and Darcy permeability fields are publicly available as part of
PDEBench~\cite{Takamoto2022}. Stationary Darcy pressure references are computed with the solver described in~\ref{app:darcy_data}.}} The 2D Navier--Stokes data were generated by the
authors with the solver described in~\ref{app:ns_data}.
{%
Coanda uses the GCA-ROM release~\cite{pichiGcaRomRelease}, {} while Helmholtz coefficient fields use CaltechDATA~\cite{dehoop2022operatordata}; we generate the four-frequency solutions using the finite element method (FEM), following~\cite{huangOperatorLearningCode}. Sod data are generated from exact Riemann solutions.
\par}

The code is available at \mbox{\url{https://github.com/rknzhl/TEMPO}}.

\clearpage

\begingroup
\small
\bibliographystyle{elsarticle-num}
\bibliography{references}
\endgroup

\newpage
\appendix

\renewcommand{\thesection}{Appendix~\Alph{section}}
\renewcommand{\thefigure}{\Alph{section}.\arabic{figure}}
\renewcommand{\thetable}{\Alph{section}.\arabic{table}}
\numberwithin{equation}{section}
\renewcommand{\theequation}{\Alph{section}.\arabic{equation}}
\renewcommand{\theHfigure}{appendix.\Alph{section}.\arabic{figure}}
\renewcommand{\theHtable}{appendix.\Alph{section}.\arabic{table}}
\renewcommand{\theHequation}{appendix.\Alph{section}.\arabic{equation}}
\setcounter{figure}{0}
\setcounter{table}{0}
\section{Supplementary Material}\label{app}
\makeatletter
\setlength{\@fptop}{0pt}
\setlength{\@fpsep}{12pt}
\setlength{\@fpbot}{0pt plus 1fil}
\makeatother

\subsection{Proof of Proposition~\ref{prop:advantage}}\label{app:proof}

\begin{lemma}\label{lem:mean}
Let $\bar{s}=\tfrac{1}{N}\sum_i s^{(i)}$ be the global mean and
$\bar{s}_m=\tfrac{1}{N_m}\sum_i\gamma_{im}s^{(i)}$ the regime-$m$ mean
with $N_m=\sum_i\gamma_{im}$, and ${d_m}=\bar{s}_m-\bar{s}$.
{Empty regimes ($N_m=0$) are omitted.}

For any rank-$K$  {$W$-orthogonal projector} $P$ and any regime $m$,
\[
  \sum_i \gamma_{im}\|(I{-}P)(s^{(i)}{-}\bar{s}_m)\|_w^2
  =
  \sum_i \gamma_{im}\|(I{-}P)(s^{(i)}{-}\bar{s})\|_w^2
  - N_m\|(I{-}P){d_m}\|_w^2.
\]
\end{lemma}
\begin{proof}
Writing $(I{-}P)(s^{(i)}{-}\bar{s}_m)=(I{-}P)(s^{(i)}{-}\bar{s})-(I{-}P){d_m}$
and expanding $\|a-b\|_w^2=\|a\|_w^2-2\langle a,b\rangle_w+\|b\|_w^2$,
then summing over $i$ with weights $\gamma_{im}$:
\begin{multline*}
  \sum_i\gamma_{im}\|(I{-}P)(s^{(i)}{-}\bar{s}_m)\|_w^2
  = \sum_i\gamma_{im}\|(I{-}P)(s^{(i)}{-}\bar{s})\|_w^2 \\
  -\;2\Bigl\langle(I{-}P)\!\sum_i\gamma_{im}(s^{(i)}{-}\bar{s}),\;(I{-}P){d_m}\Bigr\rangle_w
  +\;N_m\|(I{-}P){d_m}\|_w^2.
\end{multline*}
Since $\sum_i\gamma_{im}(s^{(i)}{-}\bar{s})=N_m{d_m}$, the cross-term equals
$-2N_m\|(I{-}P){d_m}\|_w^2$; together with the quadratic term this leaves
$-N_m\|(I{-}P){d_m}\|_w^2$, as claimed.
\end{proof}

\begin{proof}[Proof of Proposition~\ref{prop:advantage}]
Let $P^*$ and $P_m^*$ be the globally and regime-$m$ optimal rank-$K$ {$W$-orthogonal} projectors.
For a fixed rank-$K$ {$W$-orthogonal} projector $P$, the objective is minimized by any offset with
$(I{-}P)\mu = (I{-}P)\bar{s}$, so we take $\mu^*=\bar{s}$ and $\mu_m^*=\bar{s}_m$.
Since $P_m^*$ minimizes the regime-$m$ objective over all rank-$K$ {$W$-orthogonal} projectors,
substituting the (generally suboptimal) $P^*$ cannot decrease it:
\[
  \varepsilon_{\mathrm{TEMPO}}^K
  = \sum_m\sum_i\gamma_{im}\|(I{-}P_m^*)(s^{(i)}{-}\bar{s}_m)\|_w^2
  \leq\!
  \sum_m\sum_i\gamma_{im}\|(I{-}P^*)(s^{(i)}{-}\bar{s}_m)\|_w^2.
\]
Applying Lemma~\ref{lem:mean} to each $m$ and summing over $m$,
then using $\sum_m\gamma_{im}=1$ for each $i$:
\begin{multline*}
  \sum_m\sum_i\gamma_{im}\|(I{-}P^*)(s^{(i)}{-}\bar{s}_m)\|_w^2
  = \sum_i\|(I{-}P^*)(s^{(i)}{-}\bar{s})\|_w^2 \\
  - \underbrace{\sum_m N_m\|(I{-}P^*)(\bar{s}_m{-}\bar{s})\|_w^2}_{\geq\,0}
  \leq \sum_i\|(I{-}P^*)(s^{(i)}{-}\bar{s})\|_w^2
    = \varepsilon_{\mathrm{global}}^K.
\end{multline*}
\end{proof}

\subsection[Burgers': joint-model comparison]{{Burgers': joint-model comparison}}
\label{app:burgers_updated}

{%
Joint models in the common-test comparison share $3{,}000$ test trajectories. Global POD/NeuralPOD use $12{,}000$ training trajectories; TEMPO and DeepONet use $10{,}800$ for training and $1{,}200$ for validation. The 3- and 4-regime POD models use different training settings from the 2-regime model. Specialists and additional joint FNO/CNO retain their original test splits.
}

{%
Output grids are $256\times101$ for DeepONet specialists, $128\times128$ for CNO and $201\times1024$ otherwise. Joint DeepONet uses 100 training updates. Optimized TEMPO inference times are shown in parentheses.
}

\begingroup

\noindent
\begin{table}[p]
\centering
\footnotesize
\setlength{\tabcolsep}{2pt}
\begin{minipage}{\textwidth}
\normalsize\raggedright
{Numbers in parentheses give neural-network parameter counts; M denotes one million. Values with $\pm$ are means and standard deviations over three training runs. TEMPO(NeuralPOD) uses one training run per task.}
\end{minipage}
\par\medskip
\caption{{Relative $L^2$ test errors (\%) on Burgers'. Underlined: training viscosity. Bold: best in the common-test comparison. }}
\label{tab:revision_burgers_original}
\label{tab:results}
{
\resizebox{\textwidth}{!}{%
\begin{tabular}{lccrrrrr}
\toprule
Method & Train $\nu$ & Modes & $\nu=0.001$ & $\nu=0.1$ & $\nu=1$ & Mean & Time (ms) \\
\midrule
\multicolumn{8}{l}{\textit{Single-viscosity training (original split)}} \\[2pt]
DeepONet (0.553M) & 0.001 & -- & \underline{53.563} & 47.407 & \shortstack[r]{{38.173}} & 46.381 & $0.429$ \\
DeepONet (0.553M) & 0.1 & -- & 53.451 & \underline{47.110} & 34.918 & 45.160 & $0.429$ \\
DeepONet (0.553M) & 1.0 & -- & 54.729 & 48.300 & \underline{28.274} & 43.768 & $0.429$ \\
POD-DeepONet (0.468M) & 0.001 & $32$ & \underline{27.266} & 29.697 & 63.896 & 40.286 & $0.123$ \\
POD-DeepONet (0.468M) & 0.01 & $32$ & 27.438 & 27.377 & 61.436 & 38.750 & $0.126$ \\
POD-DeepONet (0.468M) & 0.1 & $32$ & 34.950 & \underline{4.797} & 34.416 & 24.721 & $0.119$ \\
POD-DeepONet (0.463M) & 1.0 & $14$ & 46.056 & 25.971 & \underline{2.523} & 24.850 & $0.093$ \\
NeuralPOD-DeepONet (4.337M) & 0.001 & $20$ & \underline{39.154} & 28.829 & 53.509 & 40.497 & $0.126$ \\
NeuralPOD-DeepONet (4.337M) & 0.01 & $20$ & 39.129 & 28.369 & 54.072 & 40.523 & $0.126$ \\
NeuralPOD-DeepONet (4.337M) & 0.1 & $20$ & 42.594 & \underline{22.910} & 37.138 & 34.214 & $0.126$ \\
NeuralPOD-DeepONet (4.337M) & 1.0 & $20$ & 48.282 & 31.525 & \underline{18.300} & 32.702 & $0.126$ \\
FNO ({1.193M}) & 0.001 & -- & \underline{42.617} & 52.855 & 62.295 & 52.589 & $10.410$ \\
FNO ({1.193M}) & 0.01 & -- & 45.406 & 40.140 & 49.171 & 44.906 & $10.464$ \\
FNO ({1.193M}) & 0.1 & -- & 64.956 & \underline{30.363} & 52.091 & 49.137 & $10.509$ \\
FNO ({1.193M}) & 1.0 & -- & 116.056 & 91.941 & \underline{15.824} & 74.607 & $10.582$ \\
CNO (2.007M) & 0.01 & -- & 68.658 & 95.479 & 234.354 & 132.830 & $3.382$ \\
CNO (2.007M) & 0.1 & -- & 98.296 & \underline{19.914} & 47.071 & 55.094 & $3.382$ \\
CNO (2.007M) & 1.0 & -- & 186.968 & 108.388 & \underline{12.521} & 102.626 & $3.382$ \\
\midrule
\multicolumn{8}{l}{\textit{Joint neural operators (original split)}} \\[2pt]
FNO ({1.193M}) & all & -- & 42.740 & 29.832 & 20.169 & 30.914 & $10.436$ \\
CNO (2.007M) & all & -- & 36.113 & 16.158 & 13.254 & 21.842 & $3.382$ \\
\midrule
\multicolumn{8}{l}{\textit{Joint training (common test set)}} \\[2pt]
DeepONet (0.790M) & all & -- & 50.361 & 34.348 & 22.820 & 35.843 & $0.413$ \\
POD-DeepONet (1.346M) & all & $64$ & \textbf{28.433} & 14.194 & 16.205 & 19.611 & $0.194$ \\
POD-DeepONet (0.485M) & all & $96$ & 32.856 & 13.365 & \textbf{11.886} & 19.369 & $0.214$ \\
NeuralPOD-DeepONet (6.449M) & all & $32$ & 36.843 & 16.267 & 18.447 & 23.853 & $0.131$ \\
TEMPO(POD) (1.397M) & all & $2\times32$ & 31.393 & \textbf{12.270} & 12.169 & \textbf{18.611} & \shortstack[r]{{$0.405\;(0.137)$}} \\
TEMPO(POD) (1.866M) & all & $3\times32$ & 30.699 & 14.276 & 16.650 & 20.542 & $0.437$ \\
TEMPO(POD) (2.318M) & all & $4\times16$ & 33.763 & 14.998 & 18.807 & 22.523 & $0.516$ \\
TEMPO(NeuralPOD) (2.917M) & all & $3\times16$ & 34.668 & 14.646 & 15.287 & 21.534 & $0.534\;(0.210)$ \\
TEMPO(NeuralPOD) (2.601M) & all & $3\times11$ & 36.206 & 15.342 & 15.816 & 22.455 & $0.394\;(0.196)$ \\
\bottomrule
\end{tabular}}}
\end{table}
\endgroup
{%
{In a separate generalization experiment, a 64-mode TEMPO(POD) trained at $\nu\in\{0.001,0.1,1\}$ gives $26.689\%$ mean relative $L^2$ error at $\nu=0.01$ on $1{,}000$ initial conditions, with no training or tuning at that viscosity.}
\par}

\begin{figure}[H]
\centering
\includegraphics[width=\textwidth]{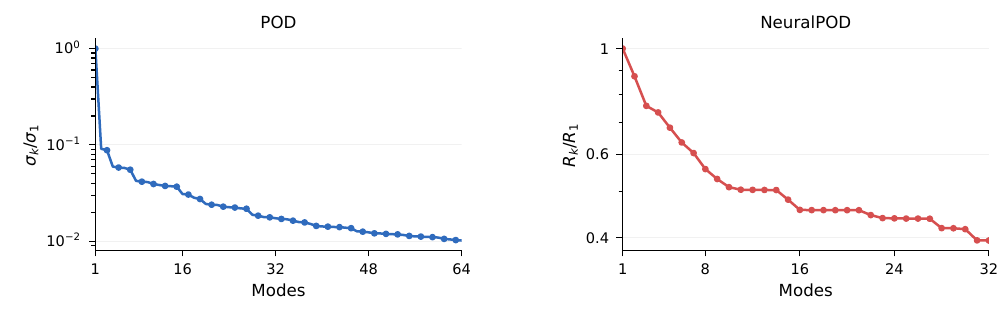}
\caption{{Burgers': normalized POD spectrum (left) and normalized squared NeuralPOD reconstruction residual (right).}}
\label{fig:basis_spectrum}
\end{figure}

\begin{figure}[!htbp]
\centering
\includegraphics[width=0.94\textwidth]{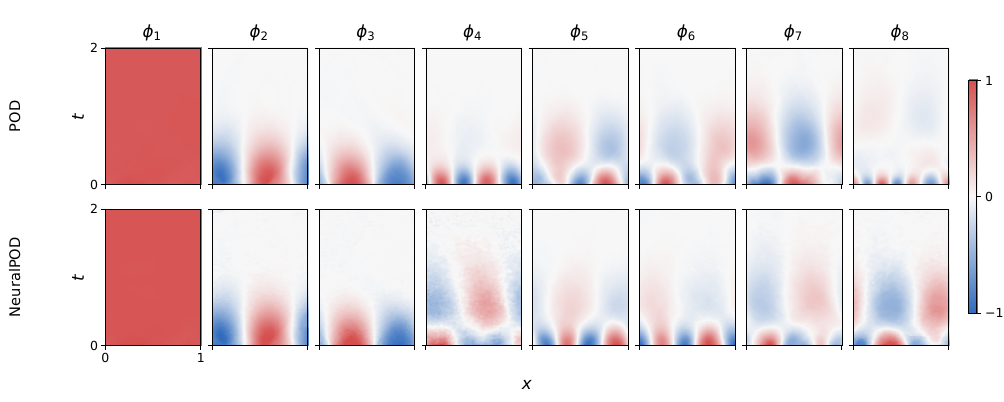}
\caption{{First eight Burgers' basis functions, scaled to unit peak magnitude: POD (top) and NeuralPOD (bottom).}}
\label{fig:basis_modes_collage}
\end{figure}

\paragraph[POD and NeuralPOD bases]{{POD and NeuralPOD bases}}
{%
Figures~\ref{fig:basis_spectrum} and~\ref{fig:basis_modes_collage} show the {} joint POD-DeepONet and Fourier NeuralPOD-DeepONet bases on the $201\times1024$ Burgers' grid. Both models use $12{,}000$ training trajectories at $\nu\in\{0.001,0.1,1\}$.
\par}

\begin{figure}[!htbp]
\centering
\includegraphics[width=0.90\textwidth]{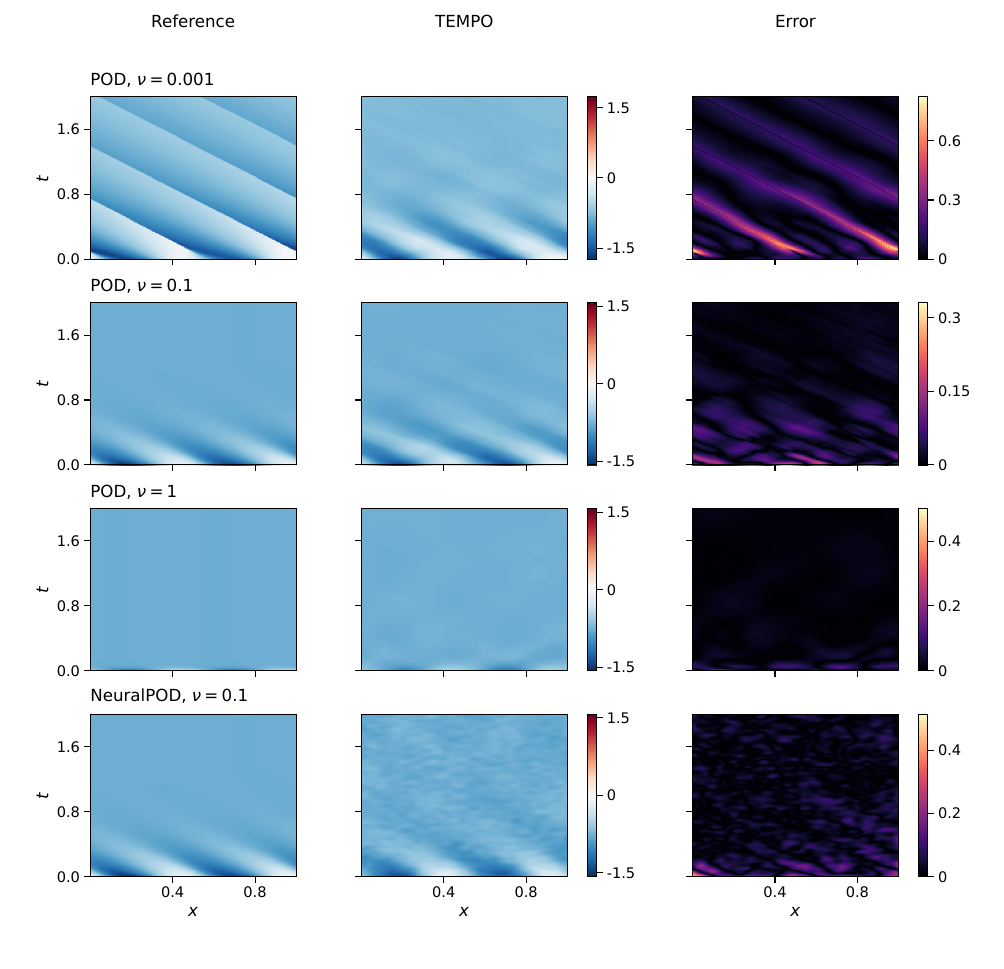}
\caption{
{{} {Burgers': reference solution, TEMPO prediction and absolute error for one initial condition. First three rows: POD at $\nu=0.001,0.1,1$. Last row: NeuralPOD at $\nu=0.1$.}}}
\label{fig:burgers_rec_collage}
\label{fig:burgers_npod_updated}
\end{figure}
\FloatBarrier

\subsection[1D Euler Data Generation]{{1D Euler Data Generation}}
\label{app:sod_data}
{%
All quantities are nondimensional. We use~\eqref{eq:sod_euler}
with $\gamma=1.4$ and piecewise-constant initial states
$(\rho_L,u_L,p_L)$ for $x<x_0$ and $(\rho_R,u_R,p_R)$ for $x\geq x_0$.
{The initial-state parameters} are sampled independently and uniformly from
\begin{align*}
 \rho_L,p_L&\in[0.85,1.15], & \rho_R&\in[0.11,0.17],\\
 u_L,u_R&\in[-0.08,0.08], & x_0&\in[0.45,0.55].
\end{align*}

{%
For each sample, we set $p_R=rp_L$ for $r\in\{0.05,0.1,0.2,0.4\}$. The inputs are $(\rho_L,u_L,p_L,\rho_R,u_R,x_0,r)$; the outputs are density, velocity and pressure.
}

{%
Exact Riemann solutions~\cite{Ketcheson2020} are sampled at 128 uniformly spaced cell centers and 16 equally spaced times up to $t=0.12$, excluding $t=0$. Waves remain inside the domain. We use $1{,}232$ training, $408$ validation and $1{,}024$ test trajectories. Initial states are disjoint across splits; all four pressure ratios for each state remain together.
}

{%
All models use the joint state $q=(\rho,u/\sqrt{1.4},p)$, with equal component and grid weights. We report its relative $L^2$ error averaged over test trajectories.
}

\par}

\begingroup

\begin{table}[!htbp]
\centering
\small
\setlength{\tabcolsep}{4pt}
\caption{{Relative $L^2$ test errors (\%) for Euler shock tubes.}}
\label{tab:revision_sod}
{
\begin{tabular}{lrlr}
\toprule
Method & Modes & Error (\%) & Time (ms) \\
\midrule
TEMPO(POD) {} {(11,866)} & $2\times24$ & \shortstack[r]{{$3.270$}} & \shortstack[r]{{$0.368\;(0.091)$}} \\
TEMPO(NeuralPOD) {(1.207M)} & $3\times16$ & \shortstack[r]{{$4.243$}} & \shortstack[r]{{$0.368\;(0.090)$}} \\
DeepONet {(0.529M)} & -- & $9.405 \pm 0.939$ & $0.106$ \\
POD-DeepONet ($K$) {(0.037M)} & $24$ & $3.989 \pm 0.002$ & $0.110$ \\
POD-DeepONet  {} {(11,952)} & $48$ & \shortstack[r]{{$3.369$}} & \shortstack[r]{{$0.157$}} \\
NeuralPOD-DeepONet {(1.168M)} & $48$ & \shortstack[r]{{$6.527$}} & \shortstack[r]{{$0.135$}} \\
FNO {(1.193M)} & -- & $5.780 \pm 0.158$ & $1.124$ \\
CNO {(2.011M)} & -- & $4.516 \pm 0.258$ & $3.101$ \\
Fixed GMM + POD {(0.109M)} & $2\times24$ & $3.400 \pm 0.064$ & $0.245$ \\
\bottomrule
\end{tabular}
}
\end{table}
\endgroup

\FloatBarrier
{A larger global POD model (427,440 parameters) gives 3.139\% test error.}

\begin{figure}[!htbp]
\centering
\includegraphics[width=\textwidth]{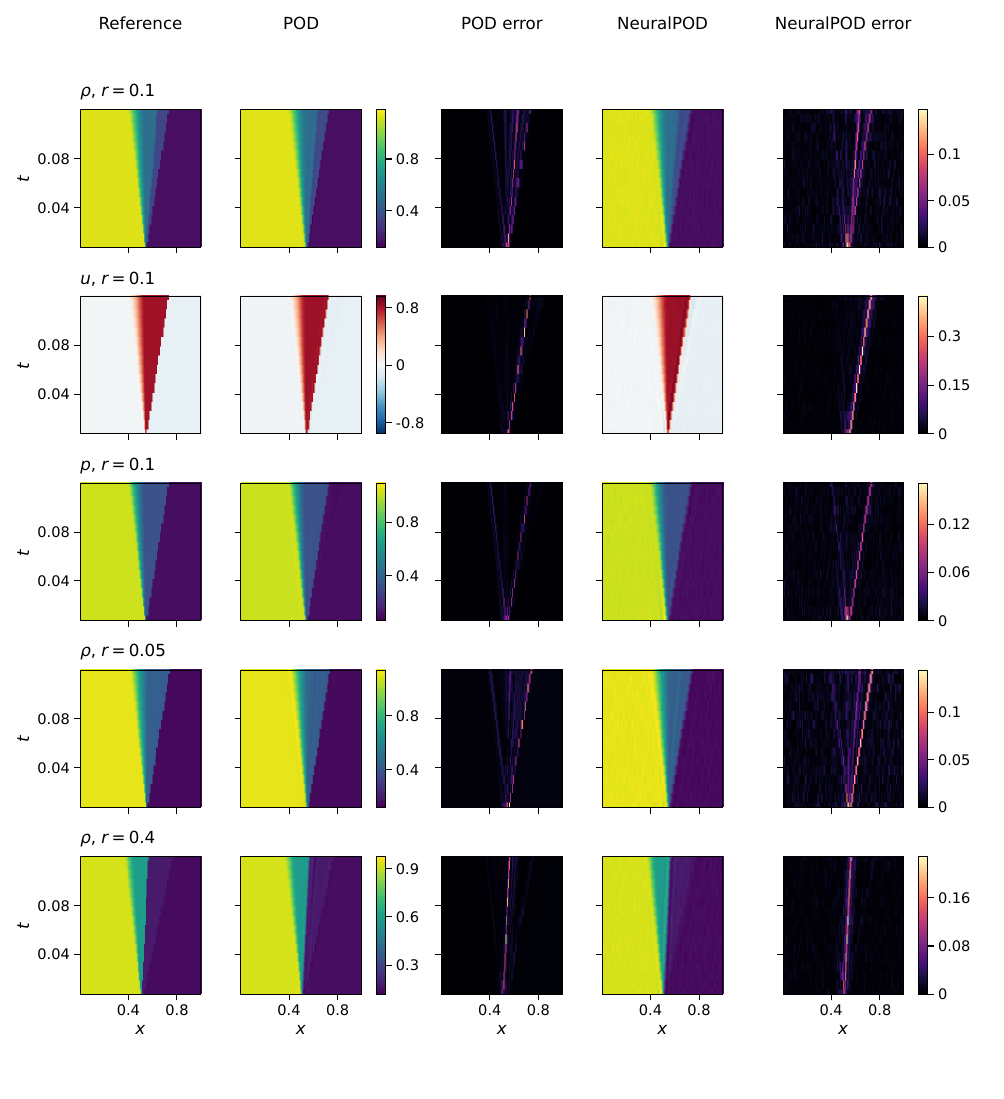}
\caption{{{} {Sod shock tubes: reference, TEMPO(POD) and TEMPO(NeuralPOD) predictions, and absolute errors. First three rows: $\rho$, $u$ and $p$ at $r=0.1$. Last two rows: $\rho$ at $r=0.05$ and $0.4$.}}}
\label{fig:sod_fields}
\label{fig:sod_parameter_fields}
\end{figure}

\subsection[Coanda flow: data and supplementary results]{{Coanda flow: data and supplementary results}}
\label{app:coanda_data}
{
The flow has parabolic inflow, no-slip walls, and the outlet condition $\nu\partial_n\mathbf{u}-p\mathbf{n}=0$.
The released parameters are ${w^{(j)}}=0.5+0.15j$, $j=0,\ldots,10$, and $\nu_\ell=2-0.05\ell$, $\ell=0,\ldots,30$~\cite{pichiGcaRomRelease}.
Training geometry indices are $\{0,1,3,5,7,9,10\}$; validation uses $\{2,8\}$ and testing $\{4,6\}$, corresponding to test half-widths $1.10$ and $1.40$.
{Velocity fields are sampled at 2,719 mesh nodes.}
Both velocity components share the original node correspondence on the $w=1.25$ reference geometry; pressure is excluded.
Inputs are standardized using training statistics, and velocities share one training {root mean square (}RMS{)} scale without centering.
FNO/CNO use reference coordinates, a fluid mask and parameters repeated on a $256\times256$ grid. Their predictions are bilinearly interpolated to the original nodes before computing the loss.
All methods are scored using the same unweighted joint nodal relative $L^2$ error, averaged over the 62 test cases.  {}
\par}

\begingroup

\begin{table}[!htbp]
\centering
\small
\setlength{\tabcolsep}{1.8pt}
\caption{{Relative $L^2$ velocity errors (\%) for Coanda flow.}}
\label{tab:revision_coanda}
{
\begin{tabular}{lrrr}
\toprule
Method & Modes & Error (\%) & Time (ms) \\
\midrule
TEMPO(POD) {(0.104M)} & $2\times12$ & $0.585 \pm 0.065$ & $0.229\;(0.065)$ \\
TEMPO(NeuralPOD) {(0.646M)} & $2\times12$ & \shortstack[r]{{$0.680$}} & \shortstack[r]{{$0.304\;(0.092)$}} \\
DeepONet {(0.528M)} & -- & $15.870 \pm 1.526$ & $0.125$ \\
POD-DeepONet ($K$) {(0.035M)} & $12$ & $1.032 \pm 0.129$ & $0.101$ \\
POD-DeepONet  {(0.107M)} & $24$ & \shortstack[r]{{$0.651$}} & \shortstack[r]{{$0.138$}} \\
NeuralPOD-DeepONet {(0.628M)} & $24$ & \shortstack[r]{{$9.117$}} & \shortstack[r]{{$0.145$}} \\
FNO {(1.193M)} & -- & $2.720 \pm 0.094$ & $2.713$ \\
CNO {(2.008M)} & -- & $2.341 \pm 0.444$ & $8.911$ \\
Fixed GMM + POD {(0.104M)} & $2\times12$ & $0.597 \pm 0.029$ & $0.299$ \\
\bottomrule\end{tabular}
}
\end{table}
\endgroup

\begin{figure}[H]
\centering
\includegraphics[width=\textwidth]{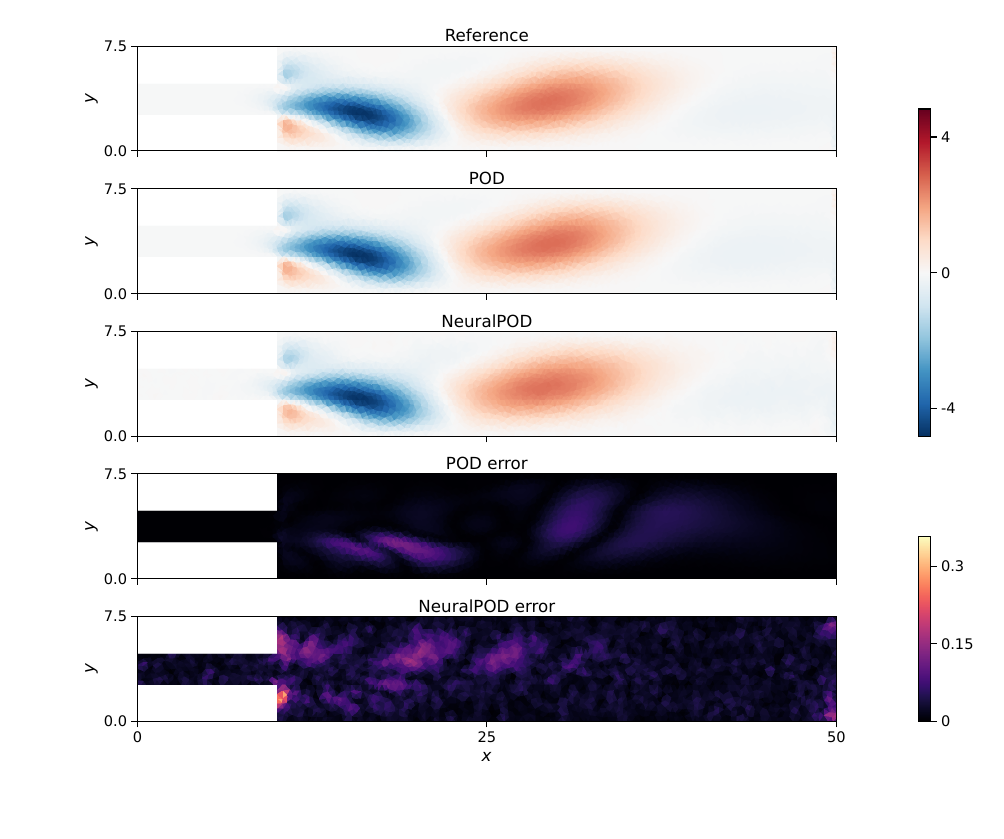}
\caption{{Coanda flow: reference, TEMPO(POD) and TEMPO(NeuralPOD) transverse velocities, followed by their absolute errors.}}
\label{fig:coanda_pod}
\label{fig:coanda_neuralpod}
\end{figure}

{%
\subsection{2D Darcy flow: data and supplementary results}
\label{app:darcy_data}
We remove exact duplicate permeability fields before splitting the data.
{Distinct fields are split into $7{,}000$ training, $1{,}000$ validation and $1{,}000$ test cases.}
Unit-source pressures are computed in float64 on $128\times128$ uniform
cells, using harmonic face permeabilities and zero Dirichlet boundary
conditions. The same pressure references and splits are used for all
models. {}

{} Reduced models use $1{,}024$ input features obtained by averaging disjoint
$4\times4$ permeability blocks; DeepONet, FNO and CNO use the full
$128\times128$ permeability field. All models predict pressure at all
$16{,}384$ cells. {} Both TEMPO variants use two $32$-mode bases. {} Their EM initialization uses a
GMM fitted to global POD coefficient vectors normalized to unit
Euclidean norm, without $\beta$.

\begin{table}[H]

\centering
\small
\setlength{\tabcolsep}{4pt}
\caption{{Relative $L^2$ pressure errors (\%) on stationary Darcy flow.}}
\label{tab:revision_darcy}
\begin{tabular}{lrr}
\toprule
Method & Modes & Test (\%) \\
\midrule
DeepONet (2.197M) & -- & 5.697 \\
FNO (1.193M) & -- & 0.628 \\
CNO (2.007M) & -- & \textbf{0.397} \\
POD-DeepONet (1.346M) & 64 & 2.254 \\
NeuralPOD-DeepONet (3.234M) & 64 & 2.280 \\
TEMPO(POD) ({} 1.343M) & $2\times32$ & {} 2.528 \\
TEMPO(NeuralPOD) (3.314M) & $2\times32$ & 2.950 \\
Fixed GMM + POD (1.396M) & $2\times32$ & 2.767 \\
\bottomrule
\end{tabular}
\par\smallskip
{\footnotesize{}}
\end{table}

\textbf{}

\begin{figure}[H]

\centering
\includegraphics[width=\textwidth]{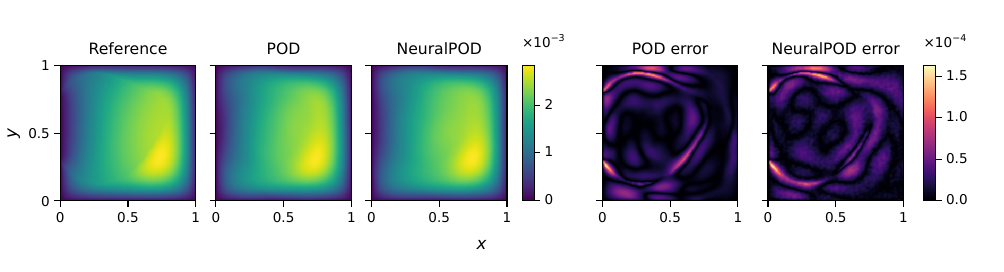}
\caption{{{} Darcy pressure at $\beta=0.01$: reference, TEMPO(POD), TEMPO(NeuralPOD) and their absolute errors.
}}
\label{fig:darcy_fields}
\end{figure}

\par}

\subsection{2D Navier--Stokes Data Generation}\label{app:ns_data}

 {We discretize Eq.~\eqref{eq:ns} with a}
pseudo-spectral solver in stream-function formulation: diffusion treated
implicitly (Crank--Nicolson), advection explicitly (second-order
Adams--Bashforth), and $\tfrac{2}{3}$-rule dealiasing on the $64{\times}64$
grid. Simulation time-step $h{=}0.01$; each recorded snapshot
aggregates $10$ sub-steps ($\Delta t{=}0.1$). Initial vorticity fields are
random sums of sinusoids,
\[
  \omega_0(\mathbf{x}) = \sum_{k=1}^{7}\frac{a_k}{1{+}k}
  \bigl[\sin({k\,x_1}+\phi_k)+\cos({k\,x_2}+\phi_k)\bigr],
\]
with $a_k\sim\mathcal{N}(0,1)$ and $\phi_k\sim\mathcal{U}[0,2\pi]$.
\ {A burn-in of $5$ time units} (500 simulation steps) discards transient
behavior before recording begins. {The first saved field is at $t=0$. The viscosities for the four Reynolds labels are $10^{-2}$, $10^{-3}$, $2.78\times10^{-4}$ and $10^{-4}$, respectively. Spatial coordinates supplied to the models are normalized as $\boldsymbol{\xi}=\mathbf{x}/(2\pi)$.}
{TEMPO uses 128 additional validation trajectories per Re.}

{%
Both TEMPO variants use four 64-mode spaces and four hidden layers of width 128. {Global POD uses 256 modes.} The inverse coefficient standardization is folded into the last branch layers for inference.

\par}
{Specialists are trained on one Reynolds number, joint models on all four. The dagger marks initialization from $\log\mathrm{Re}$.}
\par\medskip
\begin{table}[!htbp]
\centering
\footnotesize
\setlength{\tabcolsep}{3pt}
\caption{{Relative $L^2$ test errors on 2D Navier--Stokes. Underlined: in-distribution; bold: best joint model per column.}}
\label{tab:ns}
\begin{tabular}{l c c c c c r}
\toprule
Method & Train~Re & \multicolumn{4}{c}{Test~Re} & {Infer.~(ms)} \\
\cmidrule(lr){3-6}
  & & $100$ & $1000$ & $3600$ & $10000$ & \\
\midrule
\multicolumn{7}{l}{\textit{Per-Re specialists (single-Re training)}} \\[2pt]
\multirow{4}{*}{\shortstack[l]{DeepONet~\cite{Lu2019}\\{(2.691M)}}}
  & 100   & \underline{1.000} & 1.002 & 1.003 & \shortstack[r]{{1.003}} & \shortstack[r]{{0.381}} \\
  & 1000  & 1.000 & \underline{1.000} & 1.000 & 1.000 & {0.381} \\
  & 3600  & \shortstack[r]{{1.001}} & 1.000 & \underline{1.000} & 1.000 & {0.381} \\
  & 10000 & 1.001 & 1.000 & 1.000 & \underline{1.000} & {0.381} \\[4pt]
\multirow{3}{*}{\shortstack[l]{POD-DeepONet~\cite{Lu2022}\\{(2.316M)}}}
  & 100   & \underline{0.082} & 0.748 & 0.865 & 0.877 &  {0.195} \\
  & 1000  & 0.667 & \underline{0.129} & 0.671 & 0.812 & {0.163} \\
  & 3600  & 0.660 & 0.392 & \underline{0.180} & 0.623 & {0.159} \\
\shortstack[l]{POD-DeepONet~\cite{Lu2022}\\{(2.346M)}}
  & 10000 & 0.934 & 0.687 & 0.556 & \underline{0.345} & {0.273} \\[4pt]
\shortstack[l]{NeuralPOD-DeepONet\\{(14.093M)}}
  & 1000  & \shortstack[r]{{0.682}} & \shortstack[r]{{\underline{0.210}}} & \shortstack[r]{{0.567}} & \shortstack[r]{{0.756}} & \shortstack[r]{{0.154}} \\[4pt]
\multirow{4}{*}{\shortstack[l]{FNO~\cite{Li2021}\\{(2.809M)}}}
  & 100   & \underline{0.013} & 0.262 & 0.368 & 0.437 &  {1.417} \\
  & 1000  & 0.287 & \underline{0.033} & 0.138 & 0.227 & {1.397} \\
  & 3600  & 0.381 & 0.150 & \underline{0.039} & 0.128 & {1.465} \\
  & 10000 & 0.461 & 0.260 & 0.126 & \underline{0.042} & {1.340} \\[4pt]
\multirow{4}{*}{\shortstack[l]{CNO~\cite{Raonic2023}\\{(0.339M)}}}
  & 100   & \underline{0.014} & 0.047 & 0.055 & 0.058 &  {1.348} \\
  & 1000  & 0.027 & \underline{0.028} & 0.033 & 0.035 & {1.185} \\
  & 3600  & 0.030 & 0.029 & \underline{0.033} & 0.035 & {1.150} \\
  & 10000 & 0.030 & 0.028 & 0.032 & \underline{0.034} & {1.295} \\
\midrule
\multicolumn{7}{l}{\textit{Joint training (all Re)}} \\[2pt]
DeepONet~\cite{Lu2019}
  & all & \shortstack[r]{{1.000}} & \shortstack[r]{{1.000}} & \shortstack[r]{{1.000}} & \shortstack[r]{{1.000}} & \shortstack[r]{{0.380}} \\
\shortstack[l]{POD-DeepONet~\cite{Lu2022}\\{(5.486M)}}
  & all & \shortstack[r]{{0.143}} & \shortstack[r]{{0.178}} & \shortstack[r]{{0.186}} & \shortstack[r]{{0.185}} & \shortstack[r]{{0.373}} \\
FNO~\cite{Li2021} {(2.809M)}
  & all & \textbf{0.009} & \textbf{0.019} & \textbf{0.024} & \textbf{0.027} &  {1.460} \\
\shortstack[l]{NeuralPOD-DeepONet\\{(8.284M)}}
  & all & 0.700 & 0.713 & 0.314 & 0.320 & \shortstack[r]{{0.119}} \\
CNO~\cite{Raonic2023} {(0.113M)}
  & all & 0.020 & 0.035 & 0.040 & 0.043 &  {0.896} \\
\shortstack[l]{TEMPO(POD)$^\dagger$~\textit{(ours)}\\{(5.525M)}}
  & $M{=}4$ & \shortstack[r]{{0.055}} & \shortstack[r]{{0.103}} & \shortstack[r]{{0.111}} & \shortstack[r]{{0.116}} & \shortstack[r]{{0.616\;(0.340)}} \\
\shortstack[l]{TEMPO(NeuralPOD)$^\dagger$~\textit{(ours)}\\{(13.080M)}}
  & $M{=}4$ & \shortstack[r]{{0.051}} & \shortstack[r]{{0.096}} & \shortstack[r]{{0.104}} & \shortstack[r]{{0.108}} & \shortstack[r]{{0.673\;(0.460)}} \\
\bottomrule
\end{tabular}
\end{table}

\begin{figure}[H]
\centering
\includegraphics[width=\textwidth]{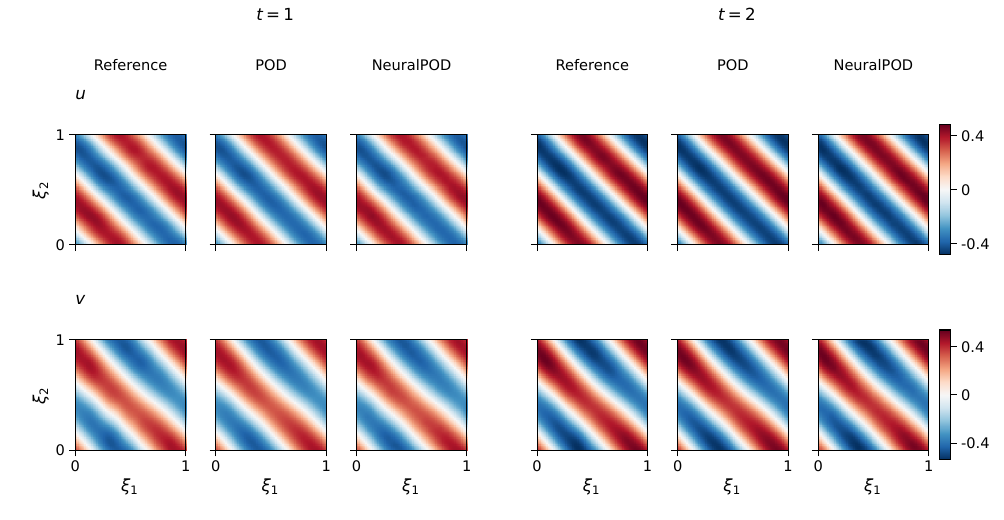}
\caption{
{{} {Navier--Stokes flow at $\mathrm{Re}=100$: reference, TEMPO(POD) and TEMPO(NeuralPOD) predictions at $t=1$ (left) and $t=2$ (right). Rows show $u$ and $v$.}}}
\label{fig:ns_rec_collage}
\end{figure}

\subsection[Helmholtz equation]{{Helmholtz equation}}
\label{app:helmholtz_data}

{%
In~\eqref{eq:helmholtz}, $\partial_n u=1$ on the top-edge interval $0.35<x<0.65$ and zero elsewhere~\cite{dehoop2022cost}. Wavespeed fields are $c=20+\tanh(a)$, where $a$ is a cosine expansion with 100 Gaussian coefficients ($d=2$, $\tau=3$). Solutions use bilinear finite elements on $100\times100$ cells~\cite{huangOperatorLearningCode}.
}

{%
We use 1,000 training, 200 validation and 200 test fields, each evaluated at all four frequencies. Wavespeed and frequency are standardized using training data. MLP field inputs are also divided by 101. All outputs are divided by a single training RMS value.
}

{%
Relative $L^2$ errors use equal node weights and are averaged over 800 test solutions. TEMPO initializes regimes from frequency and uses relative reconstruction losses.
}

\begingroup

\begin{table}[!htbp]
\centering
\small
\setlength{\tabcolsep}{1.8pt}
\caption{{Relative $L^2$ test errors (\%) for Helmholtz.}}
\label{tab:revision_helmholtz4000}
{
\begin{tabular}{lrrr}
\toprule
Method & Modes & Error (\%) & Time (ms) \\
\midrule
TEMPO(POD) {(6.704M)} & \shortstack[r]{$4\times16$} & \shortstack[r]{$3.605$} & \shortstack[r]{{$0.553\;(0.095)$}} \\
TEMPO(NeuralPOD) {(8.123M)} & \shortstack[r]{$4\times16$} & \shortstack[r]{{$5.832$}} & \shortstack[r]{{$0.470\;(0.185)$}} \\
DeepONet {(3.140M)} & -- & \shortstack[r]{$119.888$} & \shortstack[r]{$0.130$} \\
POD-DeepONet ($K$) {(1.341M)} & $16$ & \shortstack[r]{$11.598$} & \shortstack[r]{$0.113$} \\
POD-DeepONet  {(6.579M)} & \shortstack[r]{$64$} & \shortstack[r]{{$14.865$}} & \shortstack[r]{{$0.154$}} \\
NeuralPOD-DeepONet {(7.935M)} & \shortstack[r]{$64$} & \shortstack[r]{{$55.165$}} & \shortstack[r]{{$0.159$}} \\
FNO {(1.193M)} & -- & \shortstack[r]{$17.249$} & \shortstack[r]{$1.687$} \\
CNO {(2.007M)} & -- & \shortstack[r]{$6.481$} & \shortstack[r]{$3.217$} \\
Fixed GMM + POD {(6.704M)} & \shortstack[r]{$4\times16$} & \shortstack[r]{$18.468$} & \shortstack[r]{$0.466$} \\
\bottomrule\end{tabular}
}
\end{table}
\endgroup

\begin{figure}[H]
\centering
\includegraphics[width=\textwidth]{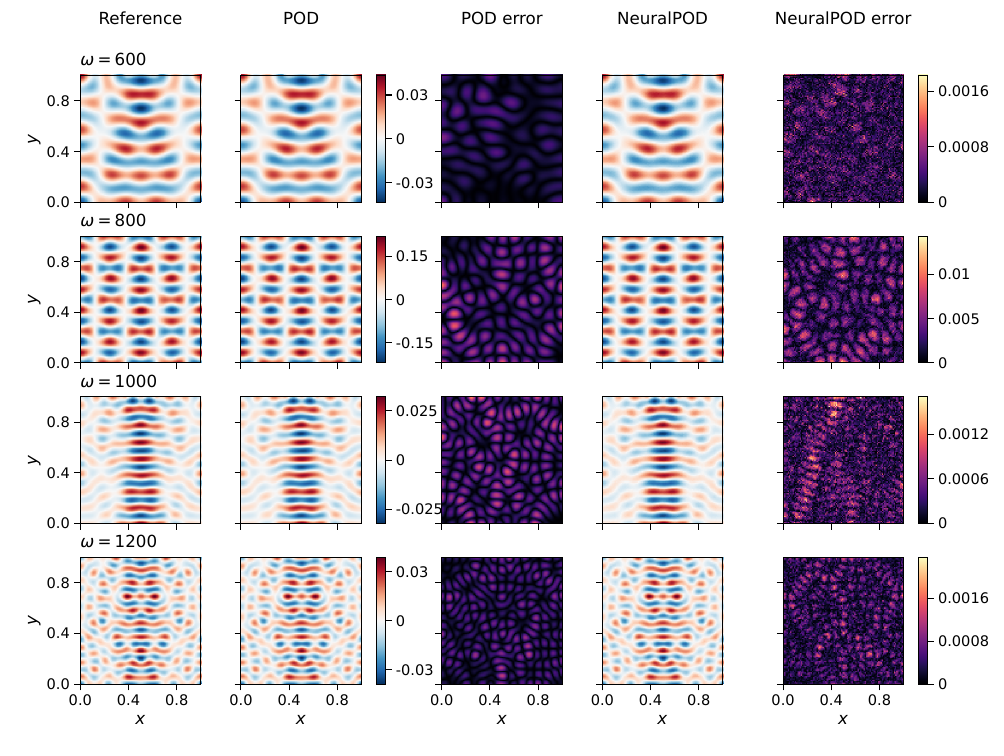}

\caption{{{} {} {Helmholtz: reference, TEMPO(POD) and TEMPO(NeuralPOD) predictions, and absolute errors for one wavespeed field at four frequencies.}}}
\label{fig:helmholtz_pod}
\label{fig:helmholtz_neuralpod}
\end{figure}

\FloatBarrier
\subsection[Model diagnostics]{{Model diagnostics}}
\label{app:frozen_diagnostics}
{%
 {} {Diagnostics compare the two TEMPO variants on the same validation trajectories, using one model per task and variant.}
\par}

\paragraph[Regime assignments and input sensitivity]{{Regime assignments and input sensitivity}}
\label{app:frozen_regimes}
{%
The two-regime Burgers' POD model has soft assignments: mean training responsibility entropy is 0.6337. For 400 validation initial conditions evaluated at all three viscosities, changing viscosity gives a mean gate $L^1$ distance of {} {0.3333}. Changing the initial condition at fixed viscosity gives mean distances of {} {0.3750--0.5433}. Thus both inputs affect routing.

For Coanda, the POD gate correlates with centerline asymmetry
(validation Spearman coefficient 0.8782).
{Asymmetry is the RMS transverse velocity on centerline nodes ($y=3.75$, $10\le x\le40$), divided by the RMS speed over all nodes.}
\par}

\paragraph[Representation and online prediction]{{Representation and online prediction}}
\label{app:frozen_reconstruction}
{%
{{For each validation trajectory, we fit coefficients in each local basis by least squares (LS) and report the lowest error.} Mixing local predictions can give lower error than using one basis alone.  These diagnostics do not separate gating and coefficient errors. SVD handles nonorthogonal or numerically dependent modes.}  {}

{%
Both variants use linear expansions in fixed modes. POD fits the modes by
weighted SVD, whereas {} {TEMPO(NeuralPOD) fits coordinate networks sequentially and refines them jointly}; the neural
parameterization does not guarantee a more accurate reduced space at a fixed mode count.

{}
{} {On Burgers', fitting coefficients across all local bases with the predicted gates fixed gives 20.48\% error, compared with 22.87\% for online prediction.} {} {} {With the predicted gating weights fixed, 80.65\% of the squared relative prediction error lies outside the combined mode span.} Coanda and Helmholtz also
have larger NeuralPOD least-squares residuals at equal mode counts
(Table~\ref{tab:frozen_reconstruction}). {} Within each local space, rescaling or an
invertible change of modes leaves its least-squares residual unchanged.
}
\par}

\begin{table}[!htbp]
\centering
\small
\setlength{\tabcolsep}{4pt}
\caption[Frozen-model reconstruction and prediction diagnostics]{{Validation errors (\%). Outside share is the fraction of squared relative prediction error outside the combined mode span (excluding mean fields).}}
\label{tab:frozen_reconstruction}
{
\begin{tabular}{l l c r r r}
\toprule
Task ($N_{\rm val}$) & Basis & Modes & \shortstack[c]{{Best local LS}} & Prediction & \shortstack[c]{{Outside share}\\{(\%)}} \\
\midrule
Burgers' (1200) & POD & $2\times32$ &14.7901&\shortstack[r]{{19.5692}}&\shortstack[r]{{54.44}}\\
 & NeuralPOD & $3\times16$ &\shortstack[r]{{23.3221}}&\shortstack[r]{{22.8662}}&\shortstack[r]{{80.65}}\\
Sod (408) & POD & $2\times24$ &{} {3.1633}&\shortstack[r]{{3.2113}}&\shortstack[r]{{88.74}}\\
 & NeuralPOD & $3\times16$ &\shortstack[r]{{4.1899}}&\shortstack[r]{{4.1909}}&\shortstack[r]{{89.06}}\\
Coanda (62) & POD & $2\times12$ &0.1396&0.8078&1.95\\
 & NeuralPOD & $2\times12$ &\shortstack[r]{{0.6915}}&\shortstack[r]{{0.8026}}&\shortstack[r]{{53.09}}\\
\shortstack[l]{Helmholtz (800)} & POD & \shortstack[r]{$4\times16$} & \shortstack[r]{$2.7967$} & \shortstack[r]{{$3.5878$}} & \shortstack[r]{{$32.11$}}\\
 & NeuralPOD & \shortstack[r]{$4\times16$} & \shortstack[r]{{4.8604}} & \shortstack[r]{{5.9033}} & \shortstack[r]{{59.80}}\\
\bottomrule
\end{tabular}}
\end{table}

\begin{samepage}
{%
We jointly optimize the local mean fields and modes, then continue training the gating and branch networks. Mode counts and data splits are unchanged.
\par}
\end{samepage}

\begin{table}[!htbp]
\centering
\small
\setlength{\tabcolsep}{10pt}
\caption{{Mean relative $L^2$ test errors (\%) of TEMPO(NeuralPOD) before and after additional training.}}
\label{tab:neuralpod_refinement}
{
\begin{tabular}{lrr}
\toprule
Task & Before & After \\
\midrule
Burgers' & 26.601 & \textbf{21.534} \\
Sod & 5.967 & \textbf{4.243} \\
Coanda & 0.698 & \textbf{0.680} \\
Navier--Stokes & 9.131 & \textbf{8.956} \\
Helmholtz & 17.984 & \textbf{5.832} \\
\bottomrule
\end{tabular}}
\end{table}

\paragraph[Inference time]{{Inference time}}
\label{app:matched_inference}

{%
For Burgers', we time the full $201\times1024$ field on an RTX A5000, one sample per call, in float32 with TF32 disabled. Basis values are cached; input normalization and field reconstruction are included. We report median times from 5 repeats of 200 individually synchronized calls after 100 warm-up calls, using 64 fixed input--parameter pairs.
}

{%
Each row of Table~\ref{tab:matched_cached_inference} compares implementations of the same model. All TEMPO experts and soft-gating terms remain active, and predictions agree up to float32 rounding. CUDA-graph timings include copying new GPU inputs and returning an independent output. The parenthesized times in Table~\ref{tab:revision_burgers_original} use additional kernel compilation for TEMPO(POD) and CUDA graphs for TEMPO(NeuralPOD).
}

{%
A separate 60-input check uses 3 repeats of 20 calls. Without cached bases, TEMPO(NeuralPOD) takes 194.27\,ms versus 426.87\,ms for global NeuralPOD; with caching, the times are 0.458 and 0.140\,ms. The global model uses larger basis networks.
}

\begin{table}[!htbp]
\centering
\small
\caption[Matched cached Burgers' inference]{{Burgers' inference times (ms) with cached bases.}}
\label{tab:matched_cached_inference}
{
\begin{tabular}{l c r r r}
\toprule
Model & Modes & Standard & Parallel heads & CUDA graph\\
\midrule
\shortstack[l]{TEMPO(POD)\\{(1.397M)}} & $2\times32$ &  {0.405} & \shortstack[r]{{0.235}} &  {0.184} \\
\shortstack[l]{Global POD\\{(1.346M)}} & 64 & \shortstack[r]{{0.194}} & \shortstack[r]{{--}} & \shortstack[r]{{0.142}} \\
\shortstack[l]{TEMPO(NeuralPOD)\\{(2.917M)}} & $3\times16$ & \shortstack[r]{{0.534}} & \shortstack[r]{{--}} & \shortstack[r]{{0.210}} \\
\shortstack[l]{Global NeuralPOD\\{(6.449M)}} & 32 & 0.131 & 0.139 & 0.104 \\
\bottomrule
\end{tabular}}
\end{table}

{%
Times in Tables~\ref{tab:revision_sod}, \ref{tab:revision_coanda} and~\ref{tab:revision_helmholtz4000} use one model per method, batch size one and an RTX A5000. {}  {Fixed-grid bases or DeepONet trunk values are cached. {} {{} {The compact Sod POD models, the TEMPO(NeuralPOD) models, the global POD controls with matched total mode counts and the global NeuralPOD controls}} use 64 training inputs, 100 warm-ups and five repeats of 200 synchronized calls, including input normalization and physical output rescaling. Other Sod and Coanda baselines use 20 warm-ups and 200 calls; other Helmholtz methods use 64 validation inputs, 100 warm-ups and five repeats of 200 calls. {} {Other TEMPO measurements use the latter repeat protocol.} Normalization and rescaling are excluded from these other measurements. Data transfers are excluded throughout. Optimized TEMPO times are in parentheses.}
\par}

{%
 Timings in Tables~\ref{tab:ns} and~\ref{tab:results} use batch size one and each model's recorded output grid. Navier--Stokes times use 100 warm-up calls and five repeats of 200 synchronized calls on 64 fixed inputs. {} {Parenthesized times use CUDA graphs for NeuralPOD and parallel branches with compiled CUDA graphs for POD; all experts remain active.}
\par}

\FloatBarrier

\begin{figure}[!htbp]
\centering
\includegraphics[width=\textwidth]{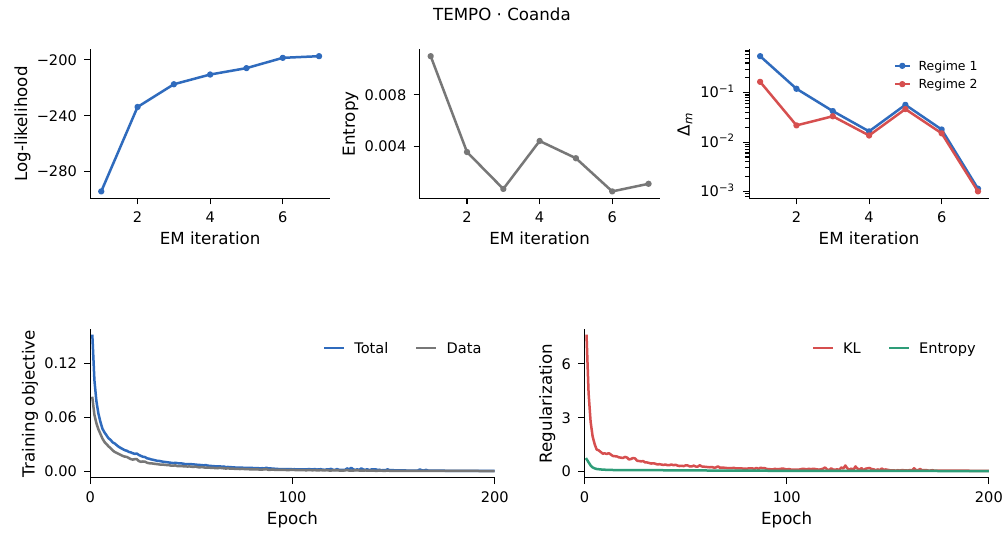}
\caption{{Offline EM convergence (top) and online training losses {(bottom, first 200 epochs)} for Coanda.}}
\label{fig:training_updated}
\end{figure}

\subsection[Auxiliary studies]{{} {Auxiliary studies}}
\label{app:archived_experiments}
\label{app:exp}
{%
These supplementary experiments use configurations distinct from the two-regime Burgers' model in Table~\ref{tab:revision_burgers_original}. 
\par}
\noindent

{Percentages count the update chosen for each regime at each EM iteration in Eq.~\eqref{eq:Delta}, with thresholds $\theta_s=0.01$ and $\theta_l=0.1$.}
\par\medskip
\begin{table}[!htbp]
\centering
\small
\setlength{\tabcolsep}{8pt}
\caption{{Adaptive basis updates on Burgers'.}}
\label{tab:adaptive}
\begin{tabular}{l l c c c c c}
\toprule
Benchmark & Basis & $M$ & EM iters & \textsc{Skip} & \textsc{Incr.} & \textsc{Full} \\
\midrule
1D Burgers' & POD       & 3 & 19 & $33\%$ & $35\%$ & $32\%$ \\
1D Burgers' & NeuralPOD & 3 & \shortstack[r]{{13}} & \shortstack[r]{{$49\%$}} & \shortstack[r]{{$13\%$}} & \shortstack[r]{{$38\%$}} \\
\bottomrule
\end{tabular}
\end{table}

{The upper block shows mean EM responsibilities and mixture weights for the 3-regime TEMPO(POD) model. The lower block shows mean predicted gating weights for the 2-regime model, evaluated on the same 400 validation initial conditions at each viscosity. For 131 of these conditions, changing viscosity changes the expert with the largest weight.}
\par\medskip
\begin{table}[!htbp]
\centering
\small
\setlength{\tabcolsep}{14pt}
\caption{{Mean regime weights on Burgers'.}}
\label{tab:gating}
\begin{tabular}{c c c c}
\toprule
$\nu$ & Regime~1 & Regime~2 & Regime~3 \\
\midrule
0.001 & 0.32 & 0.28 & 0.40 \\
0.1   & 0.32 & 0.26 & 0.42 \\
1.0   & 0.33 & 0.26 & 0.41 \\
\midrule
$\pi_m$ & 0.32 & 0.27 & 0.41 \\
\bottomrule
\end{tabular}

\par\smallskip
{
\begin{tabular}{c r r}
\toprule
$\nu$ & Mean ${g_1}$ & Mean ${g_2}$ \\
\midrule
0.001 & 0.508 & 0.492 \\
0.1   & 0.657 & 0.343 \\
1.0   & 0.754 & 0.246 \\
\bottomrule
\end{tabular}
\par\smallskip

}
\end{table}
\label{app:kappa_abl}
\begin{table}[!htbp]
\centering
\footnotesize
\setlength{\tabcolsep}{6pt}
\caption[Physical-parameter ablation]{{Increase in mean relative $L^2$ error (percentage points) when the parameter is replaced by its training mean.}}
\label{tab:kappa_abl}
\begin{tabular}{l c c}
\toprule
{Task} & {TEMPO(NeuralPOD)} & {TEMPO(POD)} \\
\midrule
{Burgers'} & {$+11.79$} & {$+28.21$} \\
{Navier--Stokes} & {$+0.70$} & {$+0.70$} \\
\bottomrule
\end{tabular}
\par\smallskip
{Burgers': 1,200 validation trajectories, $\nu=0.367$.
Navier--Stokes: 4,000 test trajectories, $\mathrm{Re}=1{,}377.45$ (mean in log space). {Models are those in Tables~\ref{tab:revision_burgers_original} and~\ref{tab:ns}.}
}
\end{table}
\FloatBarrier
\paragraph[Parameter estimation]{ {Parameter estimation}}
\label{app:viscosity_trace}

{%
For Burgers', a classifier identifies viscosity in $\{0.001,0.1,1.0\}$ from 101 measurements at one spatial point. A fixed 4-regime TEMPO(POD), with 32 modes per regime, uses the observed time series and predicted viscosity to reconstruct the observed interval. Initial conditions are disjoint across splits; validation and test conditions were not used to fit TEMPO. Classifier training duration is selected on validation. Results average 3 classifier seeds on 1,350 test trajectories; classification accuracy is 96.57\%.
}

{%
For Navier--Stokes, a linear classifier estimates $\mathrm{Re}$ from six log Fourier-band powers of the first recorded velocity field. Class probabilities are used to average $\log\mathrm{Re}$. We use 16,000 training inputs and 4,000 test trajectories, keeping both TEMPO models fixed.
}

\begin{table}[!htbp]
\centering
\small
\caption[Known and estimated physical parameters]{
{Mean relative $L^2$ errors with known and estimated physical parameters.}}
\label{tab:viscosity_trace}
\begin{tabular}{l c c}
\toprule
{Task/model} & {Known parameter} & {Estimated parameter} \\
\midrule
{Burgers', TEMPO(POD), $M=4$} & {0.32978} & {0.33283} \\
{Navier--Stokes, TEMPO(POD)} & {0.09648} & {0.09724} \\
{Navier--Stokes, TEMPO(NeuralPOD)} & {0.08956} & {0.09039} \\
\bottomrule
\end{tabular}
\end{table}

\FloatBarrier
\subsection{Algorithms}\label{app:alg}

\begin{algorithm}[H]
\caption{\textsc{NeuralBasis}}
\label{alg:npod}
\begin{algorithmic}[1]
\Require weighted snapshots $\{s^{(i)},\,\gamma_{im}\}_{i=1}^N$, tolerance $\tau$
\Ensure basis $\phi_0^{(m)},\;\bigl\{\phi_k^{(m)},\,\{\lambda_k^{(m,i)}\}_{i=1}^N\bigr\}_{k=1}^{K_m}$
\State ${\bar{s}_m} \leftarrow \displaystyle\frac{1}{N_m}\sum_{i=1}^N \gamma_{im}\,s^{(i)},\quad N_m = \sum_{i=1}^N\gamma_{im}$
  \hfill\Comment{responsibility-weighted mean trajectory}
\State Train $\phi_0^{(m)}$ by minimizing
  $\displaystyle\sum_j w_j\bigl(\phi_0^{(m)}(y_j) - {\bar{s}_{m,j}}\bigr)^2$
\State $r^{(0,i)} \leftarrow s^{(i)} - \phi_0^{(m)}(y)$
  \enspace for all $i$;\enspace $k \leftarrow 0$
\State $V_m \leftarrow \displaystyle\sum_{i=1}^N \gamma_{im}\,\|s^{(i)}-{\bar{s}_m}\|_w^2$
  \hfill\Comment{initial unmodeled variance}
\While{$\displaystyle\sum_{i=1}^N \gamma_{im}\,\|r^{(k,i)}\|_w^2
  \;\ge\; \tau\, V_m$ \textbf{ and } $k < K_{\max}$}
  \State Jointly minimize over $\phi_{k+1}^{(m)}$ and
    $\{\lambda_{k+1}^{(m,i)}\}_{i=1}^N$:
    \[
      \sum_{i=1}^N \gamma_{im}
      \sum_j w_j\!\left(\lambda_{k+1}^{(m,i)}\,\phi_{k+1}^{(m)}(y_j)
      - r^{(k,i)}_j\right)^{\!2}
    \]
  \State $r^{(k+1,i)} \leftarrow r^{(k,i)}
    - \lambda_{k+1}^{(m,i)}\,\phi_{k+1}^{(m)}(y)$
    \enspace for all $i$
  \State $k \leftarrow k + 1$
\EndWhile
\State \Return $K_m \!\leftarrow\! k$,\enspace
  $\phi_0^{(m)}$,\enspace
  $\bigl\{\phi_k^{(m)},\,\{\lambda_k^{(m,i)}\}_{i=1}^N\bigr\}_{k=1}^{K_m}$
\end{algorithmic}
\end{algorithm}

\begin{algorithm}[H]
\caption{\textsc{TEMPO}}
\label{alg:tempo}
\footnotesize
\begin{algorithmic}[1]
\Require $\mathcal{D} = \{u_0^{(i)}, \kappa^{(i)}, s^{(i)}\}_{i=1}^N$;\;
  $M,\;\theta_s,\;\theta_l,\;\theta_p,\;\theta_c,\;\tau$;\;
  {maximum EM iterations $q_{\max}$}
\Ensure Regime bases $\{\Phi_m^*\}$,\;
  amplitudes $\{\Lambda_m^*\}$,\;
  responsibilities $\{\gamma_{im}^*\}$,\;
  weights $\{\pi_m^*\}$
\Statex \textit{Initialization}
\State Compute global POD on $\{s^{(i)}\}$;
  project to $\alpha^{(i)} \in \mathbb{R}^{d_0}$ for all $i$
\State {Initialize $\gamma_{im}^{(0)}$ by GMM or the physical-parameter scheme in Section~\ref{sec:offline}.}
\State {Compute $\pi_m^{(0)}$, $a_m^{(0)}$, and $C_m^{(0)}$ from $\gamma_{im}^{(0)}$ using Eqs.~\eqref{eq:mstep_pi}, \eqref{eq:mu}, and~\eqref{eq:Sigma}.}
\For{$m = 1,\ldots,M$}
  \State $\Phi_m^{(0)},\,\Lambda_m^{(0)} \leftarrow$
    \Call{NeuralBasis}
    {$\{s^{(i)},\gamma_{im}^{(0)}\},\,\tau$}
    \hfill\Comment{or weighted SVD for TEMPO(POD), see \S\ref{sec:bases}}
\EndFor
\State $\sigma^2 \leftarrow (2N)^{-1}\!\sum_{i,m}
  \gamma_{im}^{(0)}\,\|s^{(i)}{-}\hat{s}_m^{(i)}\|_w^2$
  \hfill\eqref{eq:sigma2_calib}
\Statex \textit{EM iterations} ($q = 1, 2, \ldots$)
\Repeat
  \Statex \quad\textit{E-step}
  \For{all $i = 1,\ldots,N$ and $m = 1,\ldots,M$}
    \State Compute $\hat{s}_m^{(i)}$ via~\eqref{eq:separable}
      using $\Phi_m^{(q-1)}$ and $\Lambda_m^{(q-1)}$
    \State $\gamma_{im}^{(q)} \leftarrow$ Eq.~\eqref{eq:estep}
  \EndFor
  \Statex \quad\textit{M-step: mixture weights}
  \State $\pi_m^{(q)} \leftarrow N_m^{(q)}/N$
    for all $m$ \hfill\eqref{eq:mstep_pi}
  \Statex \quad\textit{M-step: bases}
  \For{$m = 1,\ldots,M$}
    \State Compute $a_m^{(q)}$~\eqref{eq:mu},\;
      $C_m^{(q)}$~\eqref{eq:Sigma},\;
      $\Delta_m^{(q)}$~\eqref{eq:Delta}
    \If{$\Delta_m^{(q)} < \theta_s$}
      \State $\Phi_m^{(q)},\,\Lambda_m^{(q)} \leftarrow \Phi_m^{(q-1)},\,\Lambda_m^{(q-1)}$
        \hfill \textsc{Skip}
    \ElsIf{{using POD}}
      \State {Update $\Phi_m^{(q)},\Lambda_m^{(q)}$ by weighted SVD with weights $\gamma_{im}^{(q)}$.}
    \ElsIf{$\Delta_m^{(q)} < \theta_l$}
      \hfill \textsc{Incremental}
      \State {$\bar{s}_m^{(q)} \leftarrow (N_m^{(q)})^{-1}\sum_i\gamma_{im}^{(q)}s^{(i)}$}
      \State Compute ${r_m^{(K_m,i)}}$ {for all $i$} {with the current basis};\;
        $V_m^{(q)} \leftarrow \sum_i\gamma_{im}^{(q)}\|s^{(i)}{-}\bar{s}_m^{(q)}\|_w^2$
      \If{$\sum_i\gamma_{im}^{(q)}\|r_m^{(K_m,i)}\|_w^2 \ge \tau\cdot V_m^{(q)}$ {\textbf{and} $K_m < K_{\max}$}}
        \State Train mode $K_m{+}1$ ;\; $K_m \leftarrow K_m + 1$
      \EndIf
      \If{$\|\phi_{K_m}^{(m)}\|_w^2\cdot{(N_m^{(q)})^{-1}}\displaystyle\sum_{i=1}^N
        \gamma_{im}^{(q)}\,\bigl(\lambda_{K_m}^{(m,i)}\bigr)^2 < \theta_p$}
        \State $K_m \leftarrow K_m - 1$
      \EndIf
      \State $\Phi_m^{(q)},\,\Lambda_m^{(q)} \leftarrow$ updated basis, $K_m$ modes
    \Else
      \hfill \textsc{Full rerun}
      \State $\Phi_m^{(q)},\,\Lambda_m^{(q)} \leftarrow$
        \Call{NeuralBasis}
        {$\{s^{(i)},\gamma_{im}^{(q)}\},\,\tau$}
    \EndIf
  \EndFor
\Until{$\max_m \Delta_m^{(q)} < \theta_c$ {or $q=q_{\max}$}}
\State \Return $\Phi_m^* \!\leftarrow\! \Phi_m^{(q)}$,\;
  $\Lambda_m^* \!\leftarrow\! \Lambda_m^{(q)}$,\;
  $\gamma_{im}^* \!\leftarrow\! \gamma_{im}^{(q)}$,\;
  $\pi_m^* \!\leftarrow\! \pi_m^{(q)}$
\end{algorithmic}
\end{algorithm}

\end{document}